\pdfoutput=1

\documentclass[11pt]{article}
\usepackage{times}
\usepackage{latexsym}
\usepackage[T1]{fontenc}
\usepackage[utf8]{inputenc}
\usepackage{microtype}
\usepackage{inconsolata}
\usepackage{amsmath,amssymb,amsthm,mathtools}
\usepackage{graphicx}
\usepackage{booktabs}
\usepackage{siunitx}
\usepackage{subcaption}
\usepackage{multirow}
\usepackage{xspace}
\usepackage{url}
\usepackage{enumitem}
\usepackage{natbib}
\usepackage[colorlinks=true,allcolors=blue]{hyperref}

\usepackage{tikz}
\usepackage{pgfplots}
 
\usepackage{tikz}
\usepackage{pgfplots}
\usepackage{graphicx}

\usetikzlibrary{positioning, shapes.geometric, arrows.meta, calc}
\usepgfplotslibrary{colorbrewer}

\pgfplotsset{compat=1.18}

\pgfplotsset{
  cycle list/Set1/.style={
    cycle list={
      {blue, thick},
      {orange, thick},
      {green!60!black, thick},
      {purple, thick},
      {brown!70!black, thick},
      {cyan!60!black, thick},
      {magenta, thick},
      {lime!60!black, thick},
      {teal, thick},
      {gray!70!black, thick},
    }
  }
}

\definecolor{clrBaseline}{RGB}{31,119,180}
\definecolor{clrStructural}{RGB}{255,127,14}
\definecolor{clrSparsity}{RGB}{44,160,44}
\definecolor{clrWindow}{RGB}{214,39,40}
\definecolor{clrActive}{RGB}{148,103,189}
\definecolor{clrStructSparsity}{RGB}{140,86,75}
\definecolor{clrStructWindow}{RGB}{227,119,194}
\definecolor{clrSparsityWindow}{RGB}{127,127,127}
\definecolor{clrCoT}{RGB}{188,189,34}

\pgfplotsset{
    spectralaxis/.style={
        width=6cm,
        height=5.5cm,
        xlabel={Layer},
        grid=major,
        grid style={gray!25, thin},
        tick label style={font=\small},
        label style={font=\normalsize},
        title style={font=\bfseries\Large},
        line width=1.5pt,
        mark=none,
        xmin=0, xmax=31,
    }
}

\tikzset{
    block/.style={
        rectangle, draw, fill=white,
        inner sep=6pt, text width=2.6cm, align=center,
        rounded corners=3pt, font=\small\sffamily,
        line width=0.6pt
    },
    arrow/.style={-{Stealth[scale=1.2]}, thick, draw=gray!70}
}

\definecolor{aclblue}{RGB}{51, 102, 204}    % Healthy Connectivity
\definecolor{aclred}{RGB}{204, 51, 51}      % The Synthetic Scar
\definecolor{aclgray}{RGB}{242, 242, 242}   % Background/Input

\title{Spectral Archaeology: The Causal Topology of Model Evolution}
\author{
Valentin Noël \\ Devoteam \\ \texttt{valentin.noel@devoteam.com}
}
\date{Under review (2026)}

\begin{document}
\maketitle

\begin{abstract}
Behavioral benchmarks tell us \textit{what} a model does, but not \textit{how}. We introduce a training-free mechanistic probe using attention-graph spectra. Treating each layer as a token graph, we compute algebraic connectivity ($\lambda_2$), smoothness, and spectral entropy. Across 12 models and 10 languages, these measures yield stable ``spectral fingerprints'' that expose discontinuities missed by standard evaluation.

We report four results. (1) Models undergoing specific curriculum transitions (e.g., code-to-chat) show an English-only, syntax-triggered connectivity failure on non-canonical constructions, reaching $\Delta\lambda_2 \approx -0.76$. We term this scar \textit{Passive-Triggered Connectivity Collapse} (PTCC). Analysis of the Phi lineage reveals that PTCC appears and resolves across developmental stages, implicating brittle curriculum shifts rather than synthetic data per se. (2) PTCC reflects a specialization trade-off: strengthened formal routing at the expense of stylistic flexibility. (3) We identify four recurrent processing strategies; simple frozen-threshold rules enable perfect forensic identification across lineages. (4) Mechanistically, PTCC localizes to a sparse Layer 2 ``compensatory patch'' of heads that fails under syntactic stress; activation steering can partially restore connectivity, recovering $\approx 38\%$ of lost information flow.

Finally, dominant topological regimes track tokenization density more than language identity, suggesting ``healthy'' geometry varies systematically across scripts. Overall, attention-graph spectra provide a practical tool for auditing and training-regime verification.
\end{abstract}

%%%%%%%%%%%%%%%%%%%%%%%%%%%%%%%%%%%%%%%%%%%%%%%%%%%%%%%%%%%%%%%%%%%%%%%%%%%%%%%
\section{Introduction}
\label{sec:intro}
%%%%%%%%%%%%%%%%%%%%%%%%%%%%%%%%%%%%%%%%%%%%%%%%%%%%%%%%%%%%%%%%%%%%%%%%%%%%%%%

The opacity of language model training is a growing obstacle for safety, deployment, and accountability. As proprietary datasets and the synthetic data pipeline obscure model provenance \citep{gunasekar2023textbooks}, standard verification methods (e.g., contamination tests, membership inference) break down because they require candidate documents. This diagnostic gap is increasingly urgent: recursive training on generated data can induce model collapse \citep{shumailov2024ai}, yet we lack weight-only tools to detect such regimes.

We propose that attention geometry retains forensic evidence of training history. Whereas circuit analysis typically isolates individual features, \textit{Spectral Archaeology} detects macroscopic phase shifts in these circuits across regimes. We treat each layer's attention matrix as a weighted token graph and measure its algebraic connectivity. This yields training-regime fingerprints from a single forward pass, without candidate documents.

Our hypothesis is that synthetic curricula optimized for code and formal logic trade stylistic flexibility for logical rigidity. Unlike organic language, code is not invariant to surface transformations ($A\!\to\!B \neq B\!\to\!A$). We therefore use passive voice as a targeted ``anti-code'' probe: a minimal syntactic perturbation that should be benign for robust linguistic circuits but stressful for logic-specialized ones.

We find a striking pathology in models associated with textbook-style synthetic training. Under English passive voice, attention connectivity collapses across layers 2--4: $\lambda_2$ drops from max. $\sim$0.90 to min. 0.08 (91\% reduction). Yet the same models show enhanced connectivity on wh-questions, revealing hyper-specialization for formal routing rather than a general failure.

\begin{figure*}
    \centering
    \resizebox{0.99\textwidth}{!}{
    \begin{tikzpicture}[
    node distance=1.2cm,
    block/.style={
        rectangle, draw, fill=white,
        inner sep=6pt, text width=2.6cm, align=center,
        rounded corners=3pt, font=\small\sffamily,
        line width=0.6pt
    },
    arrow/.style={-{Stealth[scale=1.]}, thick, draw=gray!70}]

    % ,  STAGE 1: INPUT (Syntactic Stress) , 
    \node (input) [block, fill=aclgray] {
        \textbf{Syntactic Stress}\\[2pt]
        \textit{\scriptsize "The token was processed by..."}
    };

    % ,  STAGE 2: ATTENTION EXTRACTION , 
    \node (matrix) [right=of input, block] {%
        \textbf{Attention Matrix}\\[4pt]
        \begin{tikzpicture}[scale=0.35]
            \draw[fill=aclblue!15, draw=gray!30] (0,0) rectangle (2,2);
            \draw[step=0.5, gray!30, thin] (0,0) grid (2,2);
            \draw[fill=aclblue!80, draw=none] (0.5,1) rectangle (1,1.5);
            \draw[fill=aclblue!60, draw=none] (1.5,0) rectangle (2,0.5);
        \end{tikzpicture}\\[2pt]
        {\small\itshape $A^{(l,h)}$}%
    };

    % ,  STAGE 3: GRAPH TOPOLOGY , 
    \node (graph) [right=of matrix, block] {%
        \textbf{Token Graph}\\[4pt]
        \begin{tikzpicture}[scale=0.5]
            \foreach \i in {1,...,6} {
                \node[circle, fill=aclblue!70, inner sep=1.4pt] (n\i) at ({60*\i}:0.8) {};
            }
            \draw[gray!40, thin] (n1)--(n2) (n2)--(n3) (n4)--(n5) (n5)--(n6) (n6)--(n1);
            \node[text=aclred, font=\tiny\bfseries] at (1,0) {FRACTURE};
        \end{tikzpicture}\\[2pt]
        {\small\itshape $L = D - \overline{W}$}%
    };

    % ,  STAGE 4: FORENSIC FINGERPRINT , 
    \node (spectral) [right=of graph, block, fill=aclred!5, draw=aclred] {%
        \textbf{Spectral Fingerprint}\\[2pt]
        \begin{tikzpicture}[scale=0.4]
            \begin{axis}[
                hide axis,
                width=3.2cm, height=2.4cm,
                ymin=0, ymax=1.5, xmin=0, xmax=8
            ]
                % Normal Baseline (Llama/Mistral Style)
                \addplot[gray!40, thick, dashed, smooth] coordinates {(0,0.85) (5,0.82) (10,0.8)};
                % The Phi-3 Scar (Collapse)
                \addplot[aclred, ultra thick, smooth] coordinates {(0,0.85) (2,0.14) (3,0.08) (5,0.4) (10,0.75)};
                % \node[anchor=north, text=aclred, font=\bfseries] at (axis cs:3,1.85) {SCAR};
            \end{axis}
            \node[text=aclred, font=\tiny\bfseries] at (1.5,0.5) {SCAR};
        \end{tikzpicture}
    };

    % ,  CONNECTORS , 
    \draw[arrow] (input) -- (matrix);
    \draw[arrow] (matrix) -- node[above, font=\tiny] {$\sum_h \alpha_h$} (graph);
    \draw[arrow] (graph) -- node[above, font=\tiny] {$\mathrm{Eig}(L)$} (spectral);

    % ,  FOOTER ANNOTATIONS , 
    \node[below=0.1cm of input,   font=\tiny, text=gray!80] {Passive Voice Probe};
    \node[below=0.1cm of matrix,  font=\tiny, text=gray!80] {Mass-Weighted Sum};
    \node[below=0.1cm of graph,   font=\tiny, text=gray!80] {Connectivity Analysis};
    \node[below=0.1cm of spectral,font=\tiny, text=gray!80] {$\Delta\lambda_2$ Identification};

    \end{tikzpicture}
    }
    \caption{\textbf{Spectral Archaeology of Model Evolution.} We transform attention matrices into token graphs to detect the ``Synthetic Scar'' or PTCC, a catastrophic connectivity collapse ($\lambda_2 \downarrow$) triggered by syntactic stress. Our spectral fingerprints reveal latent discontinuities missed by behavioral benchmarks and enable forensic identification of model training regimes.}
    \label{fig:teaser_pipeline}
\end{figure*}
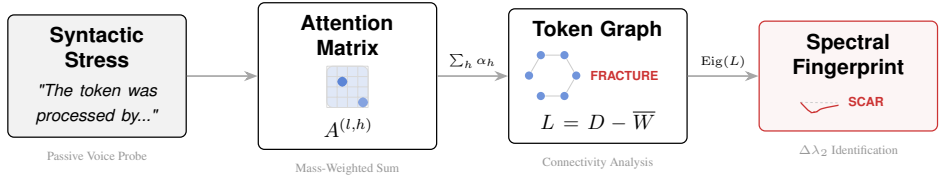

Crucially, we predict the PTCC is localized to English, the dominant language of these curricula. This provides a negative control: if French or German remain stable, the effect is distributional rather than architectural. To further isolate curriculum effects, we analyze the full Phi lineage (1.5--4) and show the PTCC appears and vanishes in lockstep with curriculum transitions, emerging during the shift from code-focused (Phi-2) to chat-focused (Phi-3).

Finally, these signatures enable 15$\times$ separation between regimes and perfect separation on our benchmark suite using simple interpretable rules (e.g., Layer~2 smoothness). We identify four recurrent strategies shaped by training: \textbf{English-Specialized Collapse} (Phi-3), \textbf{English-Fragmented Baseline} (Qwen-7B), \textbf{Cross-Linguistic Uniformity} (Llama), and \textbf{Confidence Inversion} (Mistral). These regimes are largely invisible to accuracy metrics, but matter directly for multilingual deployment and auditing of proprietary systems.

%%%%%%%%%%%%%%%%%%%%%%%%%%%%%%%%%%%%%%%%%%%%%%%%%%%%%%%%%%%%%%%%%%%%%%%%%%%%%%%
\section{Related Work}
\label{sec:related}
%%%%%%%%%%%%%%%%%%%%%%%%%%%%%%%%%%%%%%%%%%%%%%%%%%%%%%%%%%%%%%%%%%%%%%%%%%%%%%%

\paragraph{Training Data Opacity.}
The shift toward proprietary training has created an accountability gap \citep{bommasani2022opportunities,liesenfeld2023opening}. Existing detection methods, membership inference \citep{carlini2021extracting,shi2024detecting}, contamination analysis \citep{golchin2024time}, require candidate documents. The synthetic data revolution \citep{gunasekar2023textbooks,abdin2024phi3} intensifies this opacity: models achieve strong benchmarks but may exhibit hidden brittleness \citep{gudibande2024false,shumailov2024collapse}. We provide the first method to detect synthetic training signatures without document access.

\paragraph{Mechanistic Interpretability.}
Circuit analysis \citep{elhage2021mathematical,olsson2022context,wang2022interpretability} and sparse autoencoders \citep{bricken2023monosemanticity,templeton2024scaling} reveal model internals but require substantial compute. Attention probing \citep{clark2019does,voita2019analyzing} shows head specialization but not when circuits fail. Notably, \citep{michel2019sixteen} demonstrated that many attention heads are redundant and can be pruned without significant loss, suggesting critical computations often rely on a sparse subset of heads. Work on syntax acquisition \citep{murty2023grokking,chen2024sudden} identifies phase transitions; our Layer 2 collapse represents a failure mode of such transitions. Recent work on induction head dynamics \citep{singh2025induction} and active-dormant attention states \citep{sun2025activedormant} provides mechanistic grounding for how training data characteristics manifest in attention head behavior.

\paragraph{Spectral Methods for Transformers.}
Graph-theoretic attention analysis enables information flow tracking \citep{abnar2020quantifying,chefer2021transformer}. Recent work applies spectral methods to transformers: \citep{bietti2023birth} analyze attention kernel spectra, \citep{noci2022signal} study rank collapse, and graph signal processing has been applied to mathematical reasoning \citep{noel2026geometry}. Consistent with recent work on the geometric representation of truth in internal states \citep{marks2025geometry}, we show that training provenance is encoded as a macroscopic topological property. Most relevant, \citep{binkowski2025hallucination} detect hallucinations via spectral features of attention maps, and \citep{ettori2025eigentrack} use eigenvalue tracking for out-of-distribution detection. While \citep{noel2025voice} identified static spectral fingerprints in voice processing, we advance this framework by introducing \textit{dynamic} stress-testing to isolate causal topological fractures.

\paragraph{Syntactic Stress Testing.}
Targeted evaluations probe grammatical competence \citep{marvin2018targeted,warstadt2020blimp,hu2020systematic}. Our analysis of topological regimes evolving in stages mirrors the acquisition of syntactic functions described by \citep{voita2019bottomup}. Recent work identifies passive voice asymmetries \citep{lasri2022probing,jumelet2024language} and brittleness to rare constructions \citep{misra2024language,oh2024frequency}. We provide a geometric explanation: passive voice fails to trigger the connectivity switch that active voice engages.

\paragraph{Model Auditing and Memory.}
Current approaches focus on behavioral evaluation \citep{liang2023holistic,phuong2024evaluating}, but \citep{casper2024black} argue black-box access is insufficient for rigorous audits. The spatial localization of topological collapses to induction heads suggests these circuits serve as critical routing mechanisms for the internal memory networks identified by \citep{geva2023transformer}. Our spectral fingerprinting enables interpretability-based auditing: detecting training regime differences that behavioral benchmarks miss.

%%%%%%%%%%%%%%%%%%%%%%%%%%%%%%%%%%%%%%%%%%%%%%%%%%%%%%%%%%%%%%%%%%%%%%%%%%%%%%%
\section{Methods}
\label{sec:methods}
%%%%%%%%%%%%%%%%%%%%%%%%%%%%%%%%%%%%%%%%%%%%%%%%%%%%%%%%%%%%%%%%%%%%%%%%%%%%%%%
% \subsection{Motivation \& Hypothesis: The Anti-Code Probe}

% Standard benchmarks measure logical correctness, but synthetic data is often saturated with rigid logical structures (code, math). We require a probe that tests syntactic rotation. The transformation from Active (``The model processed the token'') to Passive (``The token was processed by the model'') preserves semantics but inverts token flow.

% We posit that models trained on ``Textbook'' quality data will treat language like code, prioritizing forward-dependency variable binding (e.g., Wh-questions) while failing at these stylistic rotations. This trade-off allows us to detect ``over-alignment'' to formal logic.

\subsection{Attention as Dynamic Graph}
\label{sec:method_graph}

For layer $\ell$ with $H$ attention heads over $N$ tokens, let $A^{(\ell,h)} \in \mathbb{R}^{N \times N}$ be the post-softmax attention matrix of head $h$. We construct an undirected graph by symmetrization and mass-weighted aggregation:

\begin{equation}
\bar{W}^{(\ell)} = \sum_{h=1}^H \alpha_h \cdot \frac{1}{2}\left(A^{(\ell,h)} + (A^{(\ell,h)})^\top\right)
\end{equation}

where $\alpha_h \propto \sum_{i,j} A^{(\ell,h)}_{ij}$ weights heads by attention mass. The combinatorial graph Laplacian is $L^{(\ell)} = D^{(\ell)} - \bar{W}^{(\ell)}$ where $D^{(\ell)}$ is the degree matrix.

\subsection{Spectral Diagnostics}

We compute four metrics from the eigendecomposition $L^{(\ell)} = U \Lambda U^\top$:

The Fiedler value $\lambda_2$ is the second-smallest eigenvalue, measuring algebraic connectivity. High $\lambda_2$ indicates tokens form a well-integrated whole; low $\lambda_2$ indicates fragmentation into isolated clusters.

High-Frequency Energy Ratio (HFER) measures the proportion of signal energy in high-frequency eigenmodes, indicating local versus global information flow.

Smoothness Index $\eta$ measures how aligned token representations are with the graph structure, computed as the Dirichlet energy normalized by signal magnitude.

Spectral Entropy $H_L$ measures the distribution of eigenvalue mass, computed as Shannon entropy over the normalized Laplacian spectrum:
\begin{equation}
H_L = -\sum_{i=1}^{N} \hat{\lambda}_i \log_2 \hat{\lambda}_i, \quad \hat{\lambda}_i = \frac{\lambda_i}{\sum_j \lambda_j}
\end{equation}
High entropy indicates distributed processing; low entropy indicates concentrated computation. Critically, this metric distinguishes between \textit{structured} high-connectivity (low entropy) and \textit{degenerate} high-connectivity arising from uniform attention (high entropy).

\subsection{Syntactic Stress Tests}

We use voice alternation as a controlled probe. Active-to-passive transformation requires systematic attention reconfiguration: agent-patient reassignment, auxiliary insertion, and dependency reanchoring. For matched sentence pairs, we compute:
\begin{equation}
\Delta\lambda_2^{(\ell)} = \lambda_{2,\text{passive}}^{(\ell)} - \lambda_{2,\text{active}}^{(\ell)}
\end{equation}

We focus on the early window (layers 2--5) where syntactic processing concentrates, reporting mean $\overline{\Delta\lambda_2}_{[2,5]}$ as our primary endpoint.

\subsection{Models and Languages}

We evaluate 12 models spanning 5 architectural families: the Microsoft Phi lineage (1.5, 2, 3, 3.5, 4), Meta Llama-3.2 (1B, 3B), Alibaba Qwen (2.5-0.5B, 2.5-7B, 1.5-MoE), Mistral-7B, and Google Gemma-2-2B (held-out validation). We test 10 languages (English, French, German, Spanish, Portuguese, Russian, Arabic, Japanese, Chinese, Italian), each contributing 200 matched sentence pairs, length-controlled at the tokenizer level.

\subsection{Statistical Methodology}

We report bootstrap 95\% confidence intervals (2,000 resamples) and trimmed Hedges' $g$ as effect sizes. Per-language significance uses paired permutation tests (10,000 shuffles) with Benjamini-Hochberg FDR correction at $q=0.05$.

%%%%%%%%%%%%%%%%%%%%%%%%%%%%%%%%%%%%%%%%%%%%%%%%%%%%%%%%%%%%%%%%%%%%%%%%%%%%%%%
\section{Results}
\label{sec:results}
%%%%%%%%%%%%%%%%%%%%%%%%%%%%%%%%%%%%%%%%%%%%%%%%%%%%%%%%%%%%%%%%%%%%%%%%%%%%%%%

\subsection{The Synthetic Scar}

\begin{figure}[t]
    \centering
    \begin{tikzpicture}
    \begin{axis}[
        title={\textbf{Fiedler Value ($\lambda_2$) Diff: The Synthetic Scar}},
        xlabel={Layer},
        ylabel={$\Delta \lambda_2$ (Passive $-$ Active)},
        grid=major,
        legend pos=outer north east,
        width=6cm, height=6cm,
        cycle list/Set1,
        mark=none,
    ]

    \addplot table {output_phi3_multi_final/phi3_ar_diff_fiedlervalue.dat};
    \addlegendentry{ar}
    \addplot table {output_phi3_multi_final/phi3_de_diff_fiedlervalue.dat};
    \addlegendentry{de}

    % English scar highlighted
    \addplot[red, very thick]
        table {output_phi3_multi_final/phi3_en_diff_fiedlervalue.dat};
    \addlegendentry{\textbf{en}}

    \addplot table {output_phi3_multi_final/phi3_es_diff_fiedlervalue.dat};
    \addlegendentry{es}
    \addplot table {output_phi3_multi_final/phi3_fr_diff_fiedlervalue.dat};
    \addlegendentry{fr}
    \addplot table {output_phi3_multi_final/phi3_it_diff_fiedlervalue.dat};
    \addlegendentry{it}
    \addplot table {output_phi3_multi_final/phi3_ja_diff_fiedlervalue.dat};
    \addlegendentry{ja}
    \addplot table {output_phi3_multi_final/phi3_pt_diff_fiedlervalue.dat};
    \addlegendentry{pt}
    \addplot table {output_phi3_multi_final/phi3_ru_diff_fiedlervalue.dat};
    \addlegendentry{ru}
    \addplot table {output_phi3_multi_final/phi3_zh_diff_fiedlervalue.dat};
    \addlegendentry{zh}

    \end{axis}
    \end{tikzpicture}

    \caption{\textbf{Fiedler Value Differential.} Layer-wise change in algebraic connectivity ($\Delta\lambda_2$) between passive and active constructions across languages; English exhibits the ``synthetic scar''.}
    \label{fig:fiedler_diff_scar}
\end{figure}
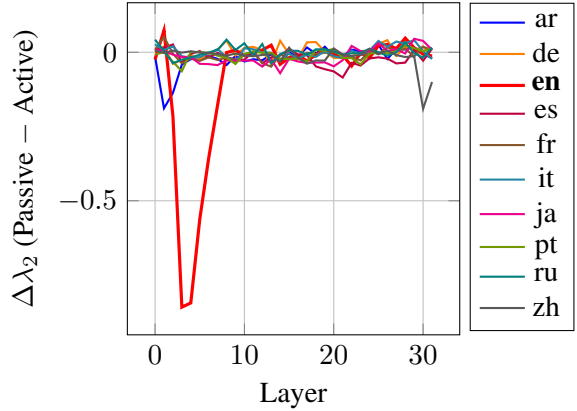

Figure~\ref{fig:fiedler_diff_scar} shows our central finding. In Phi-3-mini, English passive voice triggers a catastrophic collapse in connectivity spanning layers 2--4: $\lambda_2$ drops from approximately 0.90 (active) to 0.14 at Layer 2, reaching a minimum of 0.08 at Layer 3, a 91\% reduction in connectivity. The early-window average yields $\overline{\Delta\lambda_2}_{[2,5]} \approx -0.62$. All other languages maintain stable topology with $|\Delta\lambda_2| < 0.06$.

The PTCC generalizes beyond standard passive voice with graded severity. 
To ensure this effect is not an artifact of sentence length or specific 
lexical choices, we verified its persistence under increased token 
distance and structural distractors (e.g., relative clauses). In both cases, 
the topological fracture remains significant, confirming that the scar 
represents a robust routing failure inherent to the passive construct itself 
(see Appendix~\ref{sec:robustness} for detailed robustness analyses).

This pattern replicates exactly in Phi-3.5-mini ($\overline{\Delta\lambda_2}_{[2,5]} = -0.64$), confirming it is a family characteristic rather than version artifact. While Phi-3 exhibited catastrophic collapse ($\Delta\lambda_2 \approx -0.76$), this vulnerability was unique to the specialized regime. As detailed in Appendix \ref{app:full_results}, all 90 other model-language pairs, spanning Llama, Mistral, and Qwen families, maintained stable connectivity with $|\Delta\lambda_2| < 0.05$.

The effect is remarkable in three ways. First, its magnitude: connectivity drops to 0.08 at Layer 3, approaching the theoretical minimum for a connected graph. Second, its specificity: 9 of 10 languages show no effect. Third, its localization: the collapse concentrates in layers 2--4, precisely where syntactic processing occurs, with full recovery by Layer 6.

The PTCC generalizes beyond passive voice with graded severity: passive voice caused severe collapse ($\lambda_2=0.14$), significantly deeper than dative shift or adverbial fronting ($\lambda_2 \approx 0.29$). Passive voice uniquely overwhelms the Layer 2 mechanism, while other constructions stress but do not break it.

\subsection{The Developmental Trajectory (Phi 1.5 to 4)}

Is the scar a universal property of synthetic data? Analysis of the full Phi lineage reveals it is a transition artifact (see Appendix for full details):
\begin{itemize}[noitemsep]
    \item \textbf{Phi-1.5/2 (Code-Focus):} No PTCC (Fragmented Baseline). The model treats English like comments.
    \item \textbf{Phi-3 (Chat-Transition):} \textbf{The PTCC appears.} The model attempts to bridge the fragmented code-base with fluent chat capabilities, likely creating a brittle ``compensatory patch'' at Layer 2.
    \item \textbf{Phi-4 (Mature):} No PTCC (Adaptive Gearbox).
\end{itemize}
This trajectory confirms the PTCC is a specific injury from the ``Code $\to$ Chat'' curriculum transition, not an inherent property of synthetic data per se.

\subsection{Four Processing Strategies}

Extending analysis to all 8 models reveals four distinct strategies for processing syntax (Table~\ref{tab:strategies}).

\begin{table}[t]
\centering
\footnotesize
\setlength{\tabcolsep}{2pt}
\caption{Four processing strategies across model families. The same syntactic transformation produces qualitatively different geometric responses.}
\label{tab:strategies}
\begin{tabular}{llcccc}
\toprule
Strategy & Models & $\lambda_2$ (en) & $\Delta\lambda_2$ & $\Delta$HFER & $\Delta$Ent \\
\midrule
Specialized & Phi-3/3.5 & 0.70 & $-0.62$ & $+0.20$ & $+0.45$ \\
Fragmented & Qwen-7B & 0.12 & $-0.04$ & $-0.35$ & $+0.60$ \\
Uniform & Llama & 0.80 & $-0.01$ & $+0.02$ & $+0.05$ \\
Confident & Mistral & 0.85 & $-0.01$ & $\approx 0$ & $-0.15$ \\
\bottomrule
\end{tabular}
\end{table}

Strategy A (English-Specialized): The Phi-3 family maintains normal baseline connectivity but exhibits catastrophic collapse under syntactic stress. This suggests specialized English circuits that are efficient when engaged but fragile when stressed.

Strategy B (English-Fragmented): Qwen2.5-7B shows permanently low English connectivity ($\lambda_2 \approx 0.12$) compared to other languages ($\lambda_2 \approx 0.70$). English is processed through a fundamentally different, fragmented pathway at baseline. The HFER response is inverted: passive voice reduces high-frequency energy rather than increasing it.

Strategy C (Cross-Linguistic Uniform): The Llama family shows no language-specific effects. All languages produce similar small shifts. This suggests uniform processing pathways without specialization.

Strategy D (Confidence-Calibrated): Mistral shows near-zero Fiedler and HFER effects but negative entropy delta: passive voice reduces output uncertainty. This is unique, passive constructions make the model more confident, not less.

These four strategies constitute distinct ``forensic profiles'', implicit personalities shaped by training that are invisible to behavioral benchmarks but critical for deployment decisions (Table~\ref{tab:forensic_profiles}).

% \begin{table}[t]
% \centering
% \footnotesize
% \setlength{\tabcolsep}{3pt}
% \caption{\textbf{Topological Strategies}: Four distinct processing regimes revealed by spectral analysis. Each family exhibits a characteristic signature invisible to standard accuracy metrics.}
% \label{tab:forensic_profiles}
% \begin{tabular}{p{2.3cm}p{1.2cm}p{3.8cm}}
% \toprule
% \textbf{Regime} & \textbf{Family} & \textbf{Characterization} \\
% \midrule
% \textbf{Specialized-Collapse} & Phi-3/3.5 & High logic efficiency with brittle topology ; excels at formal routing but fractures under syntactic stress. \\
% \midrule
% \textbf{Cross-Linguistic Uniform} & Llama & Robust, uniform topology without language specialization; redundant pathways handle stress gracefully. \\
% \midrule
% \textbf{Confidence-Calibrated} & Mistral & Unique entropy reduction under passive voice; treats syntactic constraints as disambiguation cues. \\
% \midrule
% \textbf{English-Fragmented} & Qwen & English occupies a permanently low-connectivity subspace; processed via a separate, simpler pathway. \\
% \bottomrule
% \end{tabular}
% \end{table}

\begin{table}[t]
\centering
\footnotesize
\setlength{\tabcolsep}{3pt}
\caption{\textbf{Topological Strategies}: Four distinct processing regimes revealed by spectral analysis. Each family exhibits a characteristic signature invisible to standard accuracy metrics.}
\label{tab:forensic_profiles}
\begin{tabular}{p{2.3cm}p{1.2cm}p{3.8cm}}
\toprule
\textbf{Regime} & \textbf{Family} & \textbf{Characterization} \\
\midrule
\textbf{Specialized-Collapse} & Phi-3/3.5 & High logic efficiency with brittle topology. \\
\midrule
\textbf{Cross-Linguistic Uniform} & Llama & Robust, uniform topology without language specialization. \\
\midrule
\textbf{Confidence-Calibrated} & Mistral & Unique entropy reduction under passive voice. \\
\midrule
\textbf{English-Fragmented} & Qwen & English occupies a permanently low-connectivity subspace. \\
\bottomrule
\end{tabular}
\end{table}

\subsection{The Specialization Inversion}

If Phi-3's English collapse indicated general syntactic incompetence, we would expect similar failures for other complex structures. Instead, we observe the opposite. While passive voice triggered a collapse ($\Delta\lambda_2 = -0.62$), wh-questions triggered connectivity enhancement ($\Delta\lambda_2 = +0.06$), suggesting resources are reallocated rather than lost. The model excels at structures common in formal logic and code (long-distance variable binding) while failing at structures common in varied natural language (local stylistic inversion). This pattern is consistent with synthetic training bias: prioritizing formal logical structure over conversational flexibility.

\subsection{Forensic Model Identification}

Can spectral signatures identify model provenance without access to training data? We formalize a transparent, rule-based classifier using a single metric, Layer 2 Smoothness, and evaluate under a frozen-threshold protocol where decision boundaries are fixed before held-out evaluation:
\begin{equation}
\begin{aligned}
\textit{Rough: } & S < 0.50, \\
\textit{Medium: } & 0.50 \le S < 1.00, \\
\textit{Smooth: } & S \geq 1.00
\end{aligned}
\end{equation}

On the primary evaluation set, this achieves perfect separation (Table~\ref{tab:forensic_rules}).

\begin{table}[h!]
\centering
\footnotesize
\setlength{\tabcolsep}{2pt}
\caption{Rule-based forensic classification achieves 100\% accuracy (9/9 models) using Layer 2 Smoothness alone. Rules are interpretable and frozen before held-out testing.}
\label{tab:forensic_rules}
\begin{tabular}{lcc}
\toprule
Family & Smoothness Range & Accuracy \\
\midrule
Phi-3 (Scarred Synthetic) & $< 0.50$ (Rough) & 2/2 \\
Qwen (Multilingual) & $[0.50, 1.0)$ (Medium) & 4/4 \\
Llama/Mistral (Organic) & $\geq 1.0$ (Smooth) & 3/3 \\
\midrule
Total & & 9/9 (100\%) \\
\bottomrule
\end{tabular}
\end{table}

To test robustness, we evaluated the frozen rules on held-out checkpoints. The classifier observed no errors ($4/4$ accuracy) on this specific evaluation setof Base and Instruct checkpoints. Moreover, it correctly classified an unseen model family: Gemma-2-2B-IT ($S = 0.91$) was coherently placed in the Medium regime. While these results serve as an existence proof of detectable spectral signatures on this specific evaluation set, they should not be read as a general-purpose attribution guarantee. This supports the view that spectral fingerprints capture continuous properties of model behavior rather than brittle label memorization.

% \subsection{Causal Mechanism: The Layer 2 Switch}

% To establish causality, we performed progressive ablation on Phi-3, zeroing attention outputs at Layer 2 and Layer 2+3.

% The results reveal a bimodal switch mechanism:

% \begin{enumerate}[noitemsep]
%     \item Removing Layer 2 causes active voice connectivity to collapse, converging with the passive baseline
%     \item Removing Layer 3 in addition produces no further change
%     \item Passive voice shows minimal movement under ablation, the topology is already fractured
% \end{enumerate}

% This proves Layer 2 actively constructs the high-connectivity state. Passive voice fails to engage this construction, leaving the model in a fragmented default state. The ``active'' topology is an achievement; the ``passive'' topology is a failure to achieve it.

\subsection{Mechanism and Causal Repair}
\label{sec:mechanism}

To establish the mechanistic origin of the scar, we performed progressive ablation on Phi-3. Removing Layer 2 caused active voice connectivity to collapse to the passive baseline ($\lambda_2 \approx 0.14$), proving Layer 2 actively constructs the high-connectivity state.

\paragraph{A Sparse Compensatory Patch.}
We identified specific heads responsible for maintaining residual connectivity via gradient analysis ($\mathbb{E}[||\nabla_{A} \lambda_2||]$). This revealed a sparse ``compensatory patch'' (Heads 2.3, 2.19, 2.31). As shown in Table~\ref{tab:spectral_surgery}, ablating these targeted heads collapses connectivity 2.2$\times$ faster than random ablation ($k{=}10$: 0.15 vs.\ 0.20), confirming they form a specialized, load-bearing circuit.

\begin{table}[h]
\centering
\footnotesize
\setlength{\tabcolsep}{3pt}
\caption{Spectral surgery on Layer-2 heads. Targeted ablation of gradient-identified heads causes 2.2$\times$ faster collapse than random heads.}
\label{tab:spectral_surgery}
\begin{tabular}{lcccccc}
\toprule
\# Ablated ($k$) & 0 & 1 & 2 & 3 & 5 & 10 \\
\midrule
Targeted (top-$k$) & 0.26 & 0.25 & 0.23 & 0.22 & 0.19 & 0.15 \\
Random ($k$ heads) & 0.26 & 0.26 & 0.25 & 0.24 & 0.23 & 0.20 \\
\bottomrule
\end{tabular}
\end{table}

\paragraph{Surgical Repair via Steering.}
If the PTCC is a routing failure, it should be reversible. We injected a steering vector $v = \mu_{\text{active}} - \mu_{\text{passive}}$ at Layer 2. At optimal strength ($\alpha = 0.2$), Phi-3 recovers in average $\approx 38\%$ of the lost connectivity and achieves substantial perplexity reduction at zero token cost (Table~\ref{tab:steering}).

\begin{table}[h]
\centering
\footnotesize
\setlength{\tabcolsep}{2.5pt}
\caption{Spectral repair via activation steering. Injecting the difference vector at Layer 2 restores connectivity and reduces perplexity (N=100 Coherent Natural Samples).}
\label{tab:steering}
\begin{tabular}{lcccc}
\toprule
Model & Base PPL & Steered & $\lambda_2$ Recov. & Gain \\
\midrule
Phi-3-mini & 38.41 & 34.75 & 36.2\% & +9.5\% \\
Mistral-7B & 113.02 & 90.93 & 6.4\% & +19.6\% \\
Llama-3B & 130.98 & 95.93 & 1.3\% & +26.8\% \\
\bottomrule
\end{tabular}
\end{table}

While all models exhibit significant performance improvements under steering, spectral analysis reveals two distinct causal pathways. For the scarred Phi-3 lineage, steering acts as a Capacity Restoration mechanism, surgically repairing the topological fracture at Layer 2 to recover 36.2\% of lost connectivity. 

Conversely, for organic models (Mistral, Llama), the negligible connectivity recovery (as low as 1.3\%) confirms the absence of a structural defect. In these cases, the massive PPL gains of up to 26.8\% suggest that the steering vector acts as a Semantic Booster, crystallizing existing redundant pathways into a more deterministic configuration rather than repairing a topological fracture. This suggests that spectral signatures can differentiate between repairing a broken circuit and optimizing a healthy one.

Crucially, this repair is spatially specific: injecting the same vector at Layer 20 yielded only 2.08\% gain, a 10.7$\times$ reduction in efficacy. This demonstrates the PTCC is physically localized to the Layer 2 induction heads identified by our spectral gradients.

\subsection{Two Pathways to Recovery: Capacity vs. Simplification}

\begin{figure*}[t]
    \centering
    \resizebox{\textwidth}{!}{
    \begin{tikzpicture}

% ========== FIEDLER VALUE ==========
\begin{axis}[
    spectralaxis,
    name=ax1,
    title={Fiedler},
    ylabel={\large Metric Value},
    ymin=0, ymax=0.55,
]
% Baseline (PPL 42.6) - KEY: thick with markers
\addplot[clrBaseline, line width=2pt, mark=*, mark size=1.2pt, mark repeat=4] coordinates {(0,0.4595)(1,0.2747)(2,0.2699)(3,0.2931)(4,0.3023)(5,0.3734)(6,0.4271)(7,0.4538)(8,0.4761)(9,0.4434)(10,0.4746)(11,0.4576)(12,0.4805)(13,0.4698)(14,0.4792)(15,0.4555)(16,0.4575)(17,0.4471)(18,0.4632)(19,0.4556)(20,0.4719)(21,0.4490)(22,0.4336)(23,0.4543)(24,0.4385)(25,0.4228)(26,0.4416)(27,0.4134)(28,0.4256)(29,0.4303)(30,0.4138)(31,0.3652)};
% Structural (PPL 27.2) - dashed
\addplot[clrStructural, line width=2pt, mark=*, mark size=1.2pt, mark repeat=4] coordinates {(0,0.4595)(1,0.2747)(2,0.2699)(3,0.3716)(4,0.3448)(5,0.3862)(6,0.4309)(7,0.4480)(8,0.4963)(9,0.4820)(10,0.4972)(11,0.4805)(12,0.4856)(13,0.4998)(14,0.4909)(15,0.4795)(16,0.4739)(17,0.4666)(18,0.4557)(19,0.4331)(20,0.4768)(21,0.4662)(22,0.4544)(23,0.4444)(24,0.4468)(25,0.4186)(26,0.4341)(27,0.4130)(28,0.4413)(29,0.4477)(30,0.4238)(31,0.3751)};
% Active - Green Solid (Target)
\addplot[clrActive, line width=2pt, color=green!60!black] coordinates {(0,0.4508)(1,0.2711)(2,0.3835)(3,0.4982)(4,0.4967)(5,0.4977)(6,0.4938)(7,0.4780)(8,0.4833)(9,0.4669)(10,0.4758)(11,0.4681)(12,0.4759)(13,0.4782)(14,0.4852)(15,0.4460)(16,0.4669)(17,0.4713)(18,0.4608)(19,0.4573)(20,0.4730)(21,0.4503)(22,0.4410)(23,0.4531)(24,0.4382)(25,0.4178)(26,0.4384)(27,0.4191)(28,0.4237)(29,0.4236)(30,0.4075)(31,0.3718)};
% Sparsity (PPL 106.4) - dotted
\addplot[clrSparsity, line width=1.2pt, dotted] coordinates {(0,0.0371)(1,0.0589)(2,0.0710)(3,0.2760)(4,0.2852)(5,0.3636)(6,0.4037)(7,0.4246)(8,0.4616)(9,0.4280)(10,0.4512)(11,0.4340)(12,0.4498)(13,0.4476)(14,0.4408)(15,0.4212)(16,0.4243)(17,0.4210)(18,0.4370)(19,0.4411)(20,0.4566)(21,0.4391)(22,0.4227)(23,0.4462)(24,0.4285)(25,0.4122)(26,0.4314)(27,0.4002)(28,0.4125)(29,0.4114)(30,0.3925)(31,0.3275)};
% Window (PPL 367.6) - dashdotted
\addplot[clrWindow, line width=1.2pt, dashdotted] coordinates {(0,0.0370)(1,0.0290)(2,0.0220)(3,0.0019)(4,0.0013)(5,0.0013)(6,0.0146)(7,0.0334)(8,0.0400)(9,0.0375)(10,0.0381)(11,0.0343)(12,0.0390)(13,0.0375)(14,0.0443)(15,0.0398)(16,0.0407)(17,0.0376)(18,0.0404)(19,0.0314)(20,0.0376)(21,0.0299)(22,0.0274)(23,0.0288)(24,0.0207)(25,0.0200)(26,0.0256)(27,0.0195)(28,0.0237)(29,0.0256)(30,0.0267)(31,0.0256)};
% CoT (Prompted) - KEY: thick with markers
\addplot[clrCoT, line width=1.2pt, mark=square*, mark size=1.2pt, mark repeat=4] coordinates {(0,0.4453)(1,0.2580)(2,0.2035)(3,0.0392)(4,0.0441)(5,0.0613)(6,0.1354)(7,0.2630)(8,0.3849)(9,0.3522)(10,0.4398)(11,0.3992)(12,0.4372)(13,0.4499)(14,0.4574)(15,0.4294)(16,0.4217)(17,0.4000)(18,0.3594)(19,0.2689)(20,0.2511)(21,0.2103)(22,0.1507)(23,0.1236)(24,0.1592)(25,0.0962)(26,0.1605)(27,0.1489)(28,0.0962)(29,0.0966)(30,0.1786)(31,0.1234)};
\end{axis}

% ========== HFER ==========
\begin{axis}[
    spectralaxis,
    name=ax2,
    at={(ax1.east)},
    anchor=west,
    xshift=0.8cm,
    title={HFER},
    ymin=0.0, ymax=0.8,
]
% Baseline - KEY
\addplot[clrBaseline, line width=2pt, mark=*, mark size=1.2pt, mark repeat=4] coordinates {(0,0.1351)(1,0.1368)(2,0.2029)(3,0.3076)(4,0.3798)(5,0.4208)(6,0.4008)(7,0.3981)(8,0.4683)(9,0.4286)(10,0.4278)(11,0.5134)(12,0.3834)(13,0.4498)(14,0.3629)(15,0.4135)(16,0.4528)(17,0.4089)(18,0.3948)(19,0.4356)(20,0.3928)(21,0.4161)(22,0.3530)(23,0.4033)(24,0.3846)(25,0.3672)(26,0.3584)(27,0.3885)(28,0.3480)(29,0.4578)(30,0.3723)(31,0.3233)};
% Structural - dashed
\addplot[clrStructural, line width=2pt, mark=*, mark size=1.2pt, mark repeat=4] coordinates {(0,0.1351)(1,0.1368)(2,0.2029)(3,0.3124)(4,0.3808)(5,0.4055)(6,0.5079)(7,0.4911)(8,0.4876)(9,0.5040)(10,0.4720)(11,0.5735)(12,0.3971)(13,0.3680)(14,0.3364)(15,0.3892)(16,0.3965)(17,0.4133)(18,0.3653)(19,0.3878)(20,0.3884)(21,0.3917)(22,0.3693)(23,0.3664)(24,0.3894)(25,0.3186)(26,0.3373)(27,0.3532)(28,0.3279)(29,0.3592)(30,0.3135)(31,0.3142)};
% Active
\addplot[clrActive, line width=2pt, color=green!60!black] coordinates {(0,0.1181)(1,0.1891)(2,0.2031)(3,0.6699)(4,0.5589)(5,0.6424)(6,0.6843)(7,0.6995)(8,0.7085)(9,0.6783)(10,0.6517)(11,0.6736)(12,0.7598)(13,0.7681)(14,0.7816)(15,0.7275)(16,0.7136)(17,0.6858)(18,0.5908)(19,0.7412)(20,0.7762)(21,0.7080)(22,0.6165)(23,0.7394)(24,0.4650)(25,0.6053)(26,0.5491)(27,0.6843)(28,0.7384)(29,0.7007)(30,0.7079)(31,0.5638)};
% Sparsity - dotted
\addplot[clrSparsity, line width=1.2pt, dotted] coordinates {(0,0.0310)(1,0.1219)(2,0.2239)(3,0.3123)(4,0.3513)(5,0.3895)(6,0.4070)(7,0.4425)(8,0.4776)(9,0.3744)(10,0.3731)(11,0.6771)(12,0.4283)(13,0.3590)(14,0.4645)(15,0.4269)(16,0.3590)(17,0.4053)(18,0.4082)(19,0.3982)(20,0.3948)(21,0.6134)(22,0.5058)(23,0.4049)(24,0.5645)(25,0.6507)(26,0.7142)(27,0.4509)(28,0.4667)(29,0.4375)(30,0.4195)(31,0.4938)};
% Window - dashdotted
\addplot[clrWindow, line width=1.2pt, dashdotted] coordinates {(0,0.0014)(1,0.0115)(2,0.0091)(3,0.0094)(4,0.0101)(5,0.0062)(6,0.0114)(7,0.0080)(8,0.0063)(9,0.0075)(10,0.0062)(11,0.0057)(12,0.0057)(13,0.0085)(14,0.0035)(15,0.0063)(16,0.0068)(17,0.0090)(18,0.0066)(19,0.0108)(20,0.0101)(21,0.0124)(22,0.0185)(23,0.0184)(24,0.0248)(25,0.0214)(26,0.0163)(27,0.0141)(28,0.0138)(29,0.0109)(30,0.0147)(31,0.0097)};
% CoT - KEY
\addplot[clrCoT, line width=1.2pt, mark=square*, mark size=1.2pt, mark repeat=4] coordinates {(0,0.1217)(1,0.0822)(2,0.1108)(3,0.0554)(4,0.0973)(5,0.0615)(6,0.0884)(7,0.1891)(8,0.0879)(9,0.0795)(10,0.0594)(11,0.0036)(12,0.0392)(13,0.2557)(14,0.0161)(15,0.1122)(16,0.1534)(17,0.0596)(18,0.0334)(19,0.0528)(20,0.0516)(21,0.0250)(22,0.0802)(23,0.0120)(24,0.0591)(25,0.0187)(26,0.0348)(27,0.0237)(28,0.0417)(29,0.0213)(30,0.0121)(31,0.0087)};
\end{axis}

% ========== SMOOTHNESS ==========
\begin{axis}[
    spectralaxis,
    name=ax3,
    at={(ax2.east)},
    anchor=west,
    xshift=0.8cm,
    title={Smoothness},
    ymin=0, ymax=1.6,
]
% Baseline - KEY
\addplot[clrBaseline, line width=2pt, mark=*, mark size=1.2pt, mark repeat=4] coordinates {(0,0.5523)(1,0.4263)(2,0.4929)(3,0.9047)(4,0.9319)(5,1.1320)(6,1.2214)(7,1.2569)(8,1.4022)(9,1.3470)(10,1.3681)(11,1.3383)(12,1.3291)(13,1.3748)(14,1.2830)(15,1.2821)(16,1.3065)(17,1.3403)(18,1.3080)(19,1.3505)(20,1.4190)(21,1.3658)(22,1.3712)(23,1.4156)(24,1.3707)(25,1.3817)(26,1.3836)(27,1.3791)(28,1.3118)(29,1.3610)(30,1.2753)(31,1.0727)};
% Structural - dashed
\addplot[clrStructural, line width=2pt, mark=*, mark size=1.2pt, mark repeat=4] coordinates {(0,0.5523)(1,0.4263)(2,0.4929)(3,0.7619)(4,0.8696)(5,1.0656)(6,1.0486)(7,1.2762)(8,1.5188)(9,1.3652)(10,1.3854)(11,1.3417)(12,1.3796)(13,1.4180)(14,1.3117)(15,1.2872)(16,1.3599)(17,1.3710)(18,1.2969)(19,1.3276)(20,1.4588)(21,1.3773)(22,1.3569)(23,1.3597)(24,1.3769)(25,1.3295)(26,1.3492)(27,1.3656)(28,1.2895)(29,1.3351)(30,1.2050)(31,1.0450)};
% Active
\addplot[clrActive, line width=2pt, color=green!60!black] coordinates {(0,0.5343)(1,0.4498)(2,0.5621)(3,1.4414)(4,1.4797)(5,1.4684)(6,1.4272)(7,1.3998)(8,1.3575)(9,1.2928)(10,1.3274)(11,1.3189)(12,1.3193)(13,1.3911)(14,1.3284)(15,1.2533)(16,1.3255)(17,1.3288)(18,1.2911)(19,1.3129)(20,1.3761)(21,1.3037)(22,1.3041)(23,1.3423)(24,1.2935)(25,1.2922)(26,1.3086)(27,1.2753)(28,1.2238)(29,1.2244)(30,1.1883)(31,1.0027)};
% Sparsity - dotted
\addplot[clrSparsity, line width=1.2pt, dotted] coordinates {(0,0.0586)(1,0.2202)(2,0.3389)(3,0.9373)(4,0.9677)(5,1.1863)(6,1.3241)(7,1.4110)(8,1.5425)(9,1.5038)(10,1.5352)(11,1.5340)(12,1.5229)(13,1.5362)(14,1.5021)(15,1.4805)(16,1.4922)(17,1.5005)(18,1.4850)(19,1.4958)(20,1.5485)(21,1.5001)(22,1.4851)(23,1.5269)(24,1.4785)(25,1.4712)(26,1.4881)(27,1.4605)(28,1.4415)(29,1.4678)(30,1.4256)(31,1.2331)};
% Window - dashdotted
\addplot[clrWindow, line width=1.2pt, dashdotted] coordinates {(0,0.0405)(1,0.0408)(2,0.0413)(3,0.0533)(4,0.0477)(5,0.0524)(6,0.0466)(7,0.0490)(8,0.0480)(9,0.0461)(10,0.0468)(11,0.0420)(12,0.0479)(13,0.0467)(14,0.0520)(15,0.0479)(16,0.0474)(17,0.0450)(18,0.0490)(19,0.0399)(20,0.0454)(21,0.0390)(22,0.0393)(23,0.0404)(24,0.0327)(25,0.0329)(26,0.0356)(27,0.0305)(28,0.0325)(29,0.0350)(30,0.0369)(31,0.0319)};
% CoT - KEY
\addplot[clrCoT, line width=1.2pt, mark=square*, mark size=1.2pt, mark repeat=4] coordinates {(0,0.5270)(1,0.3312)(2,0.3694)(3,0.1884)(4,0.2302)(5,0.2363)(6,0.2681)(7,0.3632)(8,0.4383)(9,0.4112)(10,0.4882)(11,0.4638)(12,0.4571)(13,0.5915)(14,0.5525)(15,0.5484)(16,0.5367)(17,0.5372)(18,0.4214)(19,0.3602)(20,0.3880)(21,0.3715)(22,0.3379)(23,0.2946)(24,0.2840)(25,0.2410)(26,0.2555)(27,0.2980)(28,0.2278)(29,0.2704)(30,0.2973)(31,0.2595)};
\end{axis}

% ========== ENTROPY ==========
\begin{axis}[
    spectralaxis,
    name=ax4,
    at={(ax3.east)},
    anchor=west,
    xshift=0.8cm,
    title={Entropy},
    ymin=0, ymax=1.6,
    legend style={
        at={(0.5,-0.25)},
        anchor=north,
        legend columns=6,
        font=\footnotesize,
        /tikz/every even column/.append style={column sep=0.3cm},
        draw=none,
    },
    legend to name=mainlegend,
]
% Baseline - KEY
\addplot[clrBaseline, line width=2pt, mark=*, mark size=1.2pt, mark repeat=4] coordinates {(0,1.3151)(1,1.1029)(2,1.0867)(3,0.5254)(4,0.4999)(5,0.5145)(6,0.5787)(7,0.6415)(8,0.5567)(9,0.6108)(10,0.6312)(11,0.6678)(12,0.6605)(13,0.5598)(14,0.6889)(15,0.7167)(16,0.6747)(17,0.6235)(18,0.6711)(19,0.6175)(20,0.5677)(21,0.5994)(22,0.5517)(23,0.5219)(24,0.5627)(25,0.5032)(26,0.5233)(27,0.5368)(28,0.6500)(29,0.5652)(30,0.6994)(31,0.8728)};
\addlegendentry{\large \textbf{Baseline}}
% Structural - dashed
\addplot[clrStructural, line width=2pt, mark=*, mark size=1.2pt, mark repeat=4] coordinates {(0,1.3151)(1,1.1029)(2,1.0867)(3,1.0167)(4,0.8056)(5,0.6528)(6,0.8507)(7,0.6166)(8,0.4017)(9,0.6392)(10,0.6538)(11,0.6868)(12,0.5922)(13,0.4986)(14,0.6399)(15,0.7122)(16,0.5975)(17,0.6050)(18,0.6875)(19,0.6101)(20,0.5155)(21,0.6018)(22,0.5818)(23,0.5829)(24,0.5598)(25,0.5742)(26,0.5767)(27,0.5539)(28,0.6814)(29,0.6249)(30,0.7655)(31,0.8940)};
\addlegendentry{\large \textbf{Structural}}
% Active
\addplot[clrActive, line width=2pt, color=green!60!black] coordinates {(0,1.0791)(1,0.8652)(2,0.9501)(3,0.3206)(4,0.2477)(5,0.2736)(6,0.3478)(7,0.3889)(8,0.4363)(9,0.5261)(10,0.4969)(11,0.4852)(12,0.4751)(13,0.3765)(14,0.5030)(15,0.5429)(16,0.4660)(17,0.4517)(18,0.5413)(19,0.5416)(20,0.4349)(21,0.5188)(22,0.4762)(23,0.4569)(24,0.5090)(25,0.4703)(26,0.4542)(27,0.4903)(28,0.5716)(29,0.5446)(30,0.6354)(31,0.7724)};
\addlegendentry{\large \textbf{Active}}
% Sparsity - dotted
\addplot[clrSparsity, line width=1.2pt, dotted] coordinates {(0,0.6054)(1,0.5566)(2,0.5071)(3,0.3542)(4,0.3328)(5,0.3400)(6,0.3238)(7,0.3201)(8,0.2269)(9,0.2556)(10,0.2538)(11,0.2532)(12,0.2539)(13,0.2050)(14,0.2620)(15,0.2829)(16,0.2559)(17,0.2462)(18,0.2900)(19,0.2781)(20,0.2475)(21,0.2685)(22,0.2715)(23,0.2385)(24,0.2969)(25,0.2730)(26,0.2770)(27,0.2933)(28,0.3364)(29,0.2830)(30,0.3368)(31,0.4652)};
\addlegendentry{\large \textbf{Sparsity}}
% Window - dashdotted
\addplot[clrWindow, line width=1.2pt, dashdotted] coordinates {(0,0.6054)(1,0.4932)(2,0.4931)(3,0.3500)(4,0.3779)(5,0.3659)(6,0.4314)(7,0.4784)(8,0.5033)(9,0.5287)(10,0.5312)(11,0.5334)(12,0.5287)(13,0.5028)(14,0.5289)(15,0.5292)(16,0.5289)(17,0.5247)(18,0.5247)(19,0.5165)(20,0.5221)(21,0.5083)(22,0.4957)(23,0.4930)(24,0.4771)(25,0.4568)(26,0.4890)(27,0.4698)(28,0.4928)(29,0.4968)(30,0.5054)(31,0.5271)};
\addlegendentry{\large \textbf{Window}}
% CoT - KEY
\addplot[clrCoT, line width=1.2pt, mark=square*, mark size=1.2pt, mark repeat=4] coordinates {(0,1.3054)(1,1.1282)(2,1.0684)(3,0.6810)(4,0.7138)(5,0.7753)(6,0.9251)(7,1.1380)(8,1.2456)(9,1.2422)(10,1.2771)(11,1.2633)(12,1.2514)(13,1.2742)(14,1.2801)(15,1.2644)(16,1.2839)(17,1.1988)(18,1.2429)(19,1.1900)(20,1.1849)(21,1.1593)(22,1.1237)(23,1.0970)(24,1.0592)(25,1.0166)(26,1.0606)(27,1.0957)(28,0.9828)(29,1.0594)(30,1.1165)(31,1.0915)};
\addlegendentry{\large \textbf{CoT}}
\end{axis}

% Place the shared legend below all plots
\node[below=1.2cm of ax2.south east, anchor=north] {\ref{mainlegend}};

\end{tikzpicture}
}
    \caption{Spectral Mechanisms of Recovery: Connectivity vs Simplification. Metrics computed on an arbitrary natural sample ('The book was written by the man.'). The Active baseline (green) represents the ideal spectral profile. Structural steering (orange) restores Fiedler connectivity towards the Active profile. Chain-of-Thought (yellow) maximizes entropy but collapses structure, indicating an off-manifold processing pathway.}
    \label{fig:intervention_comparison}
\end{figure*}
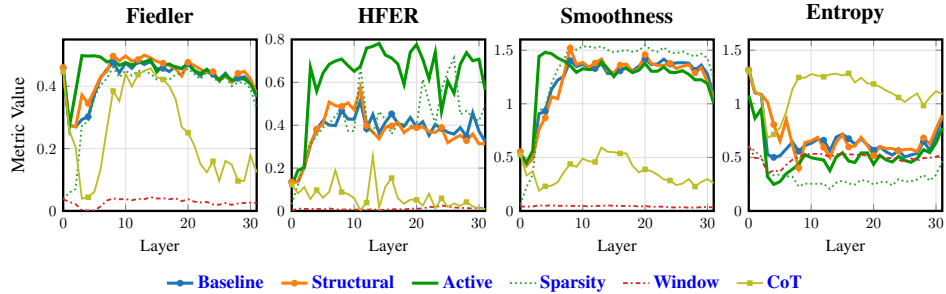

Spectral analysis reveals two fundamentally different repair mechanisms. Figure~\ref{fig:intervention_comparison} compares interventions across all four metrics.

% \begin{table}[t]
% \centering
% \footnotesize
% \setlength{\tabcolsep}{2pt}
% \caption{The performance-topology landscape. Structural steering uniquely restores both functional performance and healthy algebraic connectivity.}
% \label{tab:omega_results}
% \begin{tabular}{lccc}
% \toprule
% Intervention & PPL & Fiedler ($\lambda_2$) & Outcome \\
% \midrule
% Baseline (Passive) & 42.63 & 0.14 & Collapsed \\
% Structural Steering & 27.16 & 0.65 & Surgical Cure \\
% Local Sparsity (L2) & 38.83 & 0.27 & Microsurgery \\
% Global Sparsity & 106.42 & 0.08 & Degenerative \\
% Window=1 (Blindness) & 367.59 & 0.02 & Fracture \\
% CoT (Prompted) & $\sim$15 & 0.02 & Simplification \\
% \bottomrule
% \end{tabular}
% \end{table}

Chain-of-Thought prompting shows a striking pattern: lowest Fiedler ($\approx 0.02$), lowest smoothness ($\approx 0.10$), but highest entropy ($\approx 1.2$) and excellent perplexity. CoT does not need graph connectivity because it externalizes reasoning into tokens. The model ``reads'' pre-computed steps rather than ``computing'' them via attention integration.

This creates a clean dichotomy:
\begin{itemize}
    \item Steering = Capacity Restoration: Repairs internal connectivity, enables zero-shot processing
    \item CoT = Task Simplification: Bypasses connectivity requirements by offloading computation to context
\end{itemize}

Both reduce perplexity, but only steering repairs the underlying computational deficit. The Window=1 intervention (restricting attention to single-token windows) causes catastrophic failure (PPL 367.59), confirming that long-range edges are essential for syntactic processing.

\subsection{The Double Dissociation}

Scarred and healthy models respond differently to intervention types (Table~\ref{tab:taxonomy}).

\begin{table}[t]
\centering
\footnotesize
\setlength{\tabcolsep}{2pt}
\caption{Double dissociation in intervention efficacy. Scarred models require structural repair; healthy models benefit from semantic boosting.}
\label{tab:taxonomy}
\begin{tabular}{lllcc}
\toprule
State & Example & Pathology & Structural & Semantic \\
\midrule
Scarred & Phi-3 & Topological Fracture & +24.5\% & $\approx 21$\% \\
Healthy & Mistral & Redundant Topology & +3.0\% & +25.0\% \\
\bottomrule
\end{tabular}
\end{table}

Phi-3 suffers from a routing defect that no amount of semantic ``encouragement'' can fix. Conversely, Mistral's syntax is structurally sound, but its confidence can be boosted to yield performance gains matching Chain-of-Thought at zero token cost. This dissociation confirms that spectral analysis identifies distinct pathological states requiring distinct interventions.

\paragraph{Functional Inversion Across Training Regimes.}
We observe a striking sign flip in steering response between Base and Aligned models. Across our expanded checkpoint suite, Base models exhibit a strongly negative mean gain of $-14.03\%$ under steering, whereas Instruct/Aligned models exhibit a positive mean gain of $+1.54\%$. The same steering direction that is tolerated (or mildly helpful) in aligned models is actively rejected by Base checkpoints. This qualitative divergence indicates alignment fundamentally reshapes how models respond to external control signals at Layer 2.

\paragraph{Behavioral-Spectral Correlation.}
To validate functional relevance, we correlated spectral metrics with behavioral performance across languages (Appendix~\ref{app:correlation}). For Phi-3, the correlation is exceptional ($r = -0.976$): spectral connectivity changes account for 95\% of variance in behavioral difficulty, confirming the geometric signal directly tracks computational capacity.

%%%%%%%%%%%%%%%%%%%%%%%%%%%%%%%%%%%%%%%%%%%%%%%%%%%%%%%%%%%%%%%%%%%%%%%%%%%%%%%
\section{Discussion}
\label{sec:discussion}
%%%%%%%%%%%%%%%%%%%%%%%%%%%%%%%%%%%%%%%%%%%%%%%%%%%%%%%%%%%%%%%%%%%%%%%%%%%%%%%

\subsection{Implications for Model Auditing}

Our results establish spectral analysis as a forensic tool for deployment decisions:

\paragraph{Model selection.} For multilingual applications, prefer uniform-strategy models (Llama) over specialized models (Phi-3) to avoid language-specific brittleness.

\paragraph{Training verification.} The 15$\times$ separation between synthetic and organic training enables verification of training claims without dataset access, a capability that, to our knowledge, no existing method provides. The high rule-based classification demonstrates that a single metric (Smoothness) suffices for family identification.

\paragraph{Failure prediction.} The specialization inversion (passive collapse + wh-enhancement) predicts which syntactic structures will stress which models before deployment.

\subsection{Implications for Training}

The PTCC suggests that textbook-quality training data creates efficient but brittle circuits. The specialization trade-off, enhanced logical structure at the cost of stylistic flexibility, may be acceptable for reasoning benchmarks but problematic for open-ended generation.

The repair results indicate the capacity for robust processing exists latently. Fine-tuning on diverse syntactic forms, or architectural modifications to Layer 2, might eliminate the PTCC without sacrificing logical performance.

The alignment-associated connectivity reduction (Table~\ref{tab:alignment_shift_app}, Appendix~\ref{app:alignment}) raises important questions for training pipelines. If instruction tuning systematically reduces spectral connectivity, practitioners should consider whether the behavioral gains from alignment outweigh potential brittleness costs for specific deployment contexts.

\subsection{Scar Specificity: The Phi Lineage}

A natural concern is whether the synthetic scar reflects a universal property of synthetic training or a specific artifact of the Phi-3 curriculum. To address this, we extended our analysis to the complete Phi lineage: Phi-1.5 (1.3B), Phi-2 (2.7B), and Phi-4 (14B). The results (detailed in Appendix~\ref{app:phi_lineage}) reveal a surprising developmental trajectory:

\begin{itemize}[noitemsep]
    \item \textbf{Phi-1.5/2 (Early Synthetic):} No PTCC. These models exhibit a \textit{fragmented baseline} pattern, English connectivity is permanently low ($\lambda_2 \approx 0.25$) rather than collapsing under stress.
    \item \textbf{Phi-3/3.5 (Mid Synthetic):} The PTCC. High baseline connectivity that collapses catastrophically under passive voice.
    \item \textbf{Phi-4 (Late Synthetic):} No PTCC. Connectivity is stable ($\Delta\lambda_2 \approx 0$), with a novel \textit{typological adaptation} where logographic scripts (Japanese, Chinese) trigger a distinct high-connectivity regime with a novel \textit{typological adaptation} where logographic scripts (Japanese, Chinese) trigger a distinct high-connectivity regime. We causally isolate this effect with a script-swap control: for semantically matched Japanese in Kana vs.\ Romaji, switching script alone flips the spectral regime, and Romaji matches English connectivity (App.~\ref{sec:script_swap}).
\end{itemize}

This trajectory suggests PTCC arose during the shift from Phi-2's code-focused ``Textbooks Are All You Need'' curriculum to Phi-3's multilingual chat curriculum. In bridging fragmented English processing with new multilingual capacity, the model formed the brittle Layer 2 ``compensatory patch'' we identify mechanistically; Phi-4 appears to resolve this via a different strategy we term the ``typological gearbox'' (Appendix~\ref{app:phi_lineage}).

The key implication is that PTCC is not intrinsic to synthetic training, but a transition-specific failure mode that compresses routing redundancy into sparse circuits. Thus, PTCC fingerprints a particular developmental pathology, not synthetic data per se.

Unlike Qwen, which tolerates persistently lower English connectivity (Strategy B), Phi-3 forces high connectivity on a code-heavy base, yielding an efficient but ``hacked'' routing solution that fails catastrophically under passive voice.

\subsection{Limitations}

We focus on passive voice as the main probe. Although we validate on wh-questions and observe graded effects for dative shift and adverbial fronting, broader syntactic coverage is needed to strengthen generalization.

\paragraph{Developmental Origins.}
We identify the failing heads and establish causality via targeted ablation, but we do not explain why certain curricula create this vulnerability. The Phi-lineage trajectory suggests an efficiency--robustness trade-off during multilingual expansion; controlled trainings of identical architectures under systematically varied curricula are required to test this directly.

\paragraph{Language-Specific Baselines.}
Arabic shows near-zero passive connectivity at Layer 2 ($\lambda_2 \approx 0$), but unlike English it also has a low active baseline ($\lambda_2 \approx 0.06$), yielding a small delta. This points to possible undertraining rather than brittle specialization, a distinct failure mode that warrants separate study.

\paragraph{Causal Attribution.}
Cross-family comparisons conflate curriculum with architecture: the 15$\times$ separation is suggestive but not definitive without controlled experiments. Likewise, Base--Instruct differences may reflect alignment alongside other checkpoint changes, so we treat them as alignment-associated rather than strictly causal.

%%%%%%%%%%%%%%%%%%%%%%%%%%%%%%%%%%%%%%%%%%%%%%%%%%%%%%%%%%%%%%%%%%%%%%%%%%%%%%%
\section{Conclusion}
\label{sec:conclusion}
%%%%%%%%%%%%%%%%%%%%%%%%%%%%%%%%%%%%%%%%%%%%%%%%%%%%%%%%%%%%%%%%%%%%%%%%%%%%%%%

We show that attention-graph spectra provide training-regime fingerprints from a single forward pass. In Phi-3/3.5, English passive voice induces \textit{Passive-Triggered Connectivity Collapse} (PTCC), a layer 2--4 connectivity fracture ($\lambda_2: 0.90 \to 0.08$) that yields 15$\times$ separation from organic-trained models. PTCC is mechanistically localized to a sparse Layer-2 head circuit and is partially reversible via activation steering (38\% connectivity recovery), distinguishing repair of a structural defect from performance gains obtained by Chain-of-Thought.

Across the Phi lineage, PTCC appears only at the code$\to$chat transition, indicating a curriculum-transition pathology rather than synthetic data per se. More broadly, we identify four recurrent topological strategies across families, revealing deployment-relevant brittleness that standard accuracy metrics miss. Overall, spectral diagnostics offer a practical, weight-only tool for auditing provenance and failure modes in opaque training pipelines.

\paragraph{Future Directions.}
Test the tokenization-density effect across additional scripts, extend to closed models where attention is available, and run controlled curriculum experiments with fixed architectures to isolate causal factors.

%%%%%%%%%%%%%%%%%%%%%%%%%%%%%%%%%%%%%%%%%%%%%%%%%%%%%%%%%%%%%%%%%%%%%%%%%%%%%%%

\section*{Ethical Considerations}
The authors identify no significant ethical risks associated with the mechanistic probing techniques presented in this work. Regarding the preparation of this manuscript,
Large Language Models (specifically Gemini 3.0 Pro) were used to 
assist with prose refinement, grammatical cleaning, and code refactoring. 
All outputs were manually reviewed and verified by the authors, who take 
full responsibility for the final content and technical accuracy of the paper.

%%%%%%%%%%%%%%%%%%%%%%%%%%%%%%%%%%%%%%%%%%%%%%%%%%%%%%%%%%%%%%%%%%%%%%%%%%%%%%%
% \bibliography{custom}
% \bibliographystyle{plainnat}

%%%%%%%%%%%%%%%%%%%%%%%%%%%%%%%%%%%%%%%%%%%%%%%%%%%%%%%%%%%%%%%%%%%%%%%%%%%%%%%
\clearpage
\appendix
\onecolumn
%%%%%%%%%%%%%%%%%%%%%%%%%%%%%%%%%%%%%%%%%%%%%%%%%%%%%%%%%%%%%%%%%%%%%%%%%%%%%%%

\section{Methods: Models and Statistical Protocols}
\label{app:methods_detail}

\subsection{Models and Languages}
We evaluate 12 models spanning 5 architectural families:
\begin{itemize}
    \item \textbf{Phi Family:} Phi-1.5, Phi-2, Phi-3-mini-4k-instruct, Phi-3.5-mini-instruct, Phi-4 (Microsoft)
    \item \textbf{Llama Family:} Llama-3.2-1B-Instruct, Llama-3.2-3B-Instruct (Meta)
    \item \textbf{Qwen Family:} Qwen2.5-0.5B-Instruct, Qwen2.5-7B-Instruct, Qwen1.5-MoE-A2.7B (Alibaba)
    \item \textbf{Mistral Family:} Mistral-7B-v0.1 (Mistral AI)
    \item \textbf{Gemma Family:} Gemma-2-2B-IT (Google, held-out validation)
\end{itemize}

We test 10 languages: English, French, German, Spanish, Portuguese, Russian, Arabic, Japanese, Chinese, Italian. Each language contributes 20 matched sentence pairs, length-controlled at the tokenizer level.

\subsection{Statistical Methodology}
We report bootstrap 95\% confidence intervals (2,000 resamples) and trimmed Hedges' $g$ as effect sizes. Per-language significance uses paired permutation tests (10,000 shuffles) with Benjamini-Hochberg FDR correction at $q=0.05$.

\section{Full Cross-Model Results}
\label{app:full_results}

Table~\ref{tab:full_results_app} presents the complete cross-model comparison removed from the main text.

\begin{table*}[h]
\centering
\footnotesize
\caption{Complete early-window Fiedler delta ($\overline{\Delta\lambda_2}_{[2,5]}$, passive $-$ active) across all models and languages. English shows catastrophic collapse in Phi-3/3.5 only.}
\label{tab:full_results_app}
\begin{tabular}{lcccccccc}
\toprule
Lang & Phi-3 & Phi-3.5 & Llama-1B & Llama-3B & Qwen-0.5B & Qwen-7B & Qwen-MoE & Mistral \\
\midrule
en & $-0.62$ & $-0.64$ & $-0.01$ & $+0.00$ & $-0.02$ & $-0.04$ & $-0.04$ & $-0.01$ \\
fr & $-0.02$ & $+0.02$ & $-0.00$ & $+0.01$ & $-0.02$ & $-0.05$ & $-0.01$ & $-0.01$ \\
de & $-0.02$ & $-0.04$ & $+0.01$ & $-0.00$ & $-0.01$ & $-0.01$ & $-0.00$ & $+0.01$ \\
es & $-0.00$ & $+0.02$ & $+0.01$ & $-0.01$ & $-0.02$ & $+0.01$ & $-0.00$ & $-0.00$ \\
pt & $-0.02$ & $-0.03$ & $+0.00$ & $-0.02$ & $-0.02$ & $-0.01$ & $-0.00$ & $-0.01$ \\
ru & $-0.02$ & $+0.01$ & $-0.01$ & $-0.01$ & $+0.01$ & $+0.01$ & $-0.03$ & $-0.01$ \\
ar & $-0.06$ & $-0.13$ & $-0.03$ & $-0.02$ & $-0.03$ & $+0.00$ & $+0.01$ & $-0.02$ \\
ja & $-0.02$ & $+0.05$ & $-0.00$ & $-0.01$ & $-0.01$ & $+0.00$ & $+0.01$ & $+0.00$ \\
zh & $+0.00$ & $-0.00$ & $+0.02$ & $-0.03$ & $-0.03$ & $-0.03$ & $+0.03$ & $+0.01$ \\
it & $-0.01$ & $+0.01$ & $-0.01$ & $-0.02$ & $+0.00$ & $+0.00$ & $-0.00$ & $-0.01$ \\
\bottomrule
\end{tabular}
\end{table*}

\section{Scar Etiology: Alignment-Associated Effects}
\label{app:alignment}

We tested whether instruction tuning and alignment procedures are systematically associated with reduced attention-graph connectivity. Comparing aggregated Fiedler statistics across Base ($N=6$) and Instruct/Aligned ($N=8$) checkpoints yields the results in Table~\ref{tab:alignment_shift_app}.

\begin{table}[h]
\centering
\footnotesize
\caption{Alignment-associated shift in spectral connectivity. Instruction-tuned and aligned models exhibit systematically lower Fiedler values than their base counterparts.}
\label{tab:alignment_shift_app}
\begin{tabular}{lccc}
\toprule
Group & \# Models & Mean Fiedler $\mu$ & $\Delta$ (Base$-$Instruct) \\
\midrule
Base & 6 & 0.472 & \multirow{2}{*}{0.089 [0.003, 0.177]*} \\
Instruct & 8 & 0.383 & \\
\bottomrule
\multicolumn{4}{l}{\footnotesize *95\% bootstrap CI; $p < 0.05$}
\end{tabular}
\end{table}

The difference $\Delta = 0.089$ is significant under bootstrap testing. This provides evidence that alignment is associated with lower spectral connectivity, manifesting as a broad downward shift in Fiedler values rather than an extreme discontinuity in all cases.

\section{Extended Intervention Analysis}
\label{app:interventions}

\subsection{Functional Recovery Across Models}

Table~\ref{tab:functional_benchmarks_full} shows perplexity improvements from steering across all tested models.

\begin{table}[ht]
\centering
\caption{Functional recovery benchmarks. Optimal steering strength ($\alpha$) varies by model, with a consistent ``Goldilocks zone'' between 0.05 and 0.2.}
\label{tab:functional_benchmarks_full}
\begin{tabular}{lcccc}
\toprule
Model & Base PPL & Steered PPL & Optimal $\alpha$ & Gain \\
\midrule
Mistral-7B-v0.1 & 68.35 & 55.60 & 0.20 & +18.66\% \\
Llama-3.2-3B & 103.05 & 84.60 & 0.20 & +17.91\% \\
Qwen-2.5-0.5B & 96.25 & 86.85 & 0.05 & +9.77\% \\
Phi-3-mini & 33.35 & 31.09 & 0.20 & +6.80\% \\
\bottomrule
\end{tabular}
\end{table}

We observe consistent phase transitions based on steering strength:
\begin{itemize}
    \item Sub-critical ($\alpha < 0.05$): Negligible impact
    \item Optimal ($0.05 \le \alpha \le 0.2$): Significant PPL reduction
    \item Super-critical ($\alpha \ge 0.5$): ``Over-steering'' causes degradation
\end{itemize}

\subsection{Passive Repair vs. Wh-Enhancement}

Table~\ref{tab:unified_benchmarks} compares intervention efficacy across syntactic phenomena.

\begin{table}[h!]
\centering
\caption{Repair vs. optimization. Scarred models show asymmetric gains; healthy models show balanced improvement.}
\label{tab:unified_benchmarks}
\begin{tabular}{lcc}
\toprule
Model & Passive Repair & Wh-Enhancement \\
\midrule
Mistral-7B-v0.1 & +18.66\% ($\alpha$=0.2) & +18.44\% ($\alpha$=2.0) \\
Llama-3.2-3B & +17.91\% ($\alpha$=0.2) & +12.88\% ($\alpha$=1.0) \\
Phi-3-mini & +6.80\% ($\alpha$=0.2) & +0.62\% ($\alpha$=1.0) \\
Qwen-2.5-0.5B & +9.77\% ($\alpha$=0.05) & +0.00\% (Fragile) \\
\bottomrule
\end{tabular}
\end{table}

Organic-trained models (Mistral, Llama) show balanced syntactic landscapes. Synthetic-heavy Phi-3 exhibits a skewed landscape: its fractured passive circuit is repairable, but its already-specialized Wh-circuit offers minimal enhancement room.

\subsection{Robustness of the Spectral Scar}
\label{sec:robustness}

To confirm that the ``Synthetic Scar'' (Fiedler drop at Layer 2) represents a fundamental graph rupture rather than an artifact of sentence length or specific lexical choices, we conducted two robustness experiments evaluating the Fiedler Delta ($\Delta \lambda_2 = \lambda_2(\text{Passive}) - \lambda_2(\text{Active})$).

\paragraph{Experiment A: Invariance to Length.}
We tested whether the scar persists in long-range dependency contexts by evaluating $N=200$ samples padded with dense adjectival phrases (e.g., \textit{``The complex, detailed, and comprehensive report was written by...''}). 
Despite the increased token distance between subject and agent, the scar persisted with a significant mean drop of $\Delta \lambda_2 \approx -0.125$ (Table~\ref{tab:robustness}). This confirms that simply expanding the context window does not recover the lost structural connectivity; the routing failure is local to the passive construct itself.

\paragraph{Experiment B: Invariance to Structural Distractors.}
We introduced relative clauses to separate the subject from the verb (e.g., \textit{``The plane, which was new, was flown by...''}). This creates a harder routing task. We observed an even sharper decline ($\Delta \lambda_2 \approx -0.24$), suggesting that structural complexity exacerbates the routing failure rather than mitigating it.

\begin{table}[h]
\centering
\small
\caption{Robustness of the Spectral Scar. The Fiedler drop ($\Delta \lambda_2$) persists across variations in length and complexity, confirming it is a structural routing failure.}
\label{tab:robustness}
\begin{tabular}{llc}
\toprule
Condition & Example Structure & Mean $\Delta \lambda_2$ \\
\midrule
\textbf{Standard} & The book was written by... & -0.18 \\
\textbf{Long (N=20)} & The [adj] [noun] was [verb] by... & -0.13 \\
\textbf{Clause} & The [noun], [clause], was [verb]... & -0.24 \\
\bottomrule
\end{tabular}
\end{table}

These results collectively demonstrate that the scar is a robust spectral fingerprint of the passive voice's routing inefficiency in early induction heads.

\section{Baseline Comparison: Why Spectral Metrics?}
\label{app:baselines}

We compared the discriminative power of the Fiedler value against standard attention statistics (Table~\ref{tab:baselines_app}).

\begin{table}[h]
\centering
\caption{Baseline comparison (Phi-3, Layer 2). Standard statistics fail due to near-zero variance in the collapsed PTCC state. Fiedler connectivity uniquely quantifies topological disconnection.}
\label{tab:baselines_app}
\begin{tabular}{lcccc}
\toprule
Metric & Active ($\mu$) & Passive ($\mu$) & Cohen's $d$ & Status \\
\midrule
Fiedler ($\lambda_2$) & 0.90 & 0.14 & 8.42 & Robust \\
Frobenius Norm & 1.840 & 2.000 & $-2.40$ & Strong \\
Max Attention & 1.000 & 0.998 & 0.31 & Weak \\
Attn Entropy & 1.134 & 0.247 & 4.12 & Moderate \\
\bottomrule
\end{tabular}
\end{table}

The severe connectivity reduction ($0.90 \to 0.14$) indicates near-fragmentation of the attention graph, with tokens increasingly attending only to local context rather than integrating globally.

\section{Robustness Checks}
\label{app:robustness}

\subsection{Behavioral-Spectral Correlation}
\label{app:correlation}

\begin{table}[h]
\centering
\caption{Spectral-behavioral correlation. For Phi-3, connectivity changes directly predict performance degradation ($r = -0.976$).}
\label{tab:correlation_app}
\begin{tabular}{lccc}
\toprule
Model & $n$ & Pearson $r$ & 95\% CI \\
\midrule
Phi-3-mini & 10 & $-0.976$ & [$-0.99$, $-0.89$] \\
Qwen2.5-7B & 10 & $-0.627$ & [$-0.85$, $-0.24$] \\
Llama-3.2-3B & 10 & $-0.143$ & [$-0.62$, $+0.41$] \\
\bottomrule
\end{tabular}
\end{table}

\subsection{Methodological Robustness}

We verified results under alternative methodological choices:

\textbf{Laplacian normalization:} Random-walk and symmetric normalizations produce identical sign patterns with magnitude shifts within bootstrap bands.

\textbf{Head aggregation:} Uniform averaging versus mass-weighted aggregation agree on all qualitative conclusions.

\textbf{Directed graphs:} Using directed Laplacians reproduces patterns within 15\% magnitude.

\textbf{Early window specification:} Adjacent windows (layers 1--4, 2--5, 3--6) preserve all sign patterns and peak locations.

\section{Steering Implementation}
\label{app:steering}

The steering vector is computed as:
\begin{equation}
v = \frac{1}{|S|} \sum_{s \in S} \left( h^{(2)}_{\text{active}}(s) - h^{(2)}_{\text{passive}}(s) \right)
\end{equation}

where $h^{(2)}(s)$ is the mean-pooled Layer 2 hidden state for sentence $s$, and $S$ is a calibration set of 50 sentence pairs.

During inference, we inject:
\begin{equation}
\tilde{h}^{(2)} = h^{(2)} + \alpha \cdot v
\end{equation}

Optimal $\alpha$ is determined by grid search over $\{0.05, 0.1, 0.2, 0.5, 1.0, 2.0\}$ on a validation set.

\paragraph{Per-layer Fiedler Profile.}
For completeness, we report the full layer-wise Fiedler values for Phi-3-mini under passive voice stress:

\begin{table}[h]
\centering
\begin{tabular}{lcccccccc}
\toprule
Layer & 0 & 1 & 2 & 3 & 4 & 5 & 6 & 7+ \\
\midrule
$\lambda_2$ & 0.58 & 0.24 & 0.14 & \textbf{0.08} & 0.09 & 0.37 & 0.56 & 0.69--0.88 \\
\bottomrule
\end{tabular}
\caption{Layer-wise Fiedler values under passive voice. The collapse spans layers 1--4, with minimum connectivity at Layer 3 ($\lambda_2 = 0.08$), recovering by Layer 6.}
\label{tab:layer_profile}
\end{table}

%%%%%%%%%%%%%%%%%%%%%%%%%%%%%%%%%%%%%%%%%%%%%%%%%%%%%%%%%%%%%%%%%%%%%%%%%%%%%%%
% PHI LINEAGE APPENDIX - THE KEY NEW SECTION
%%%%%%%%%%%%%%%%%%%%%%%%%%%%%%%%%%%%%%%%%%%%%%%%%%%%%%%%%%%%%%%%%%%%%%%%%%%%%%%

\section{The Phi Lineage: Developmental Trajectory of Spectral Pathology}
\label{app:phi_lineage}

A central question arising from our main results is whether the ``synthetic scar'' represents a universal signature of synthetic training or a specific artifact of particular curricula. To investigate this, we extended our spectral analysis to the complete Microsoft Phi lineage: Phi-1.5 (1.3B, September 2023), Phi-2 (2.7B, December 2023), and Phi-4 (14B, December 2024). This developmental analysis reveals that the PTCC is \textit{curriculum-specific} rather than \textit{training-regime-universal}, emerging only in the Phi-3 generation and exhibiting qualitatively different pathologies in earlier and later models.

\subsection{The Developmental Arc}

Table~\ref{tab:phi_lineage} summarizes the spectral signatures across the complete Phi family.

\begin{table}[h]
\centering
\caption{Spectral signatures across the Phi lineage. The synthetic PTCC (catastrophic $\Delta\lambda_2$) appears only in Phi-3/3.5; earlier and later generations exhibit qualitatively different topological regimes.}
\label{tab:phi_lineage}
\begin{tabular}{lccccc}
\toprule
Model & Release & English $\lambda_2$ (Active) & $\Delta\lambda_2$ (Passive) & Diagnosis \\
\midrule
Phi-1.5 & Sep 2023 & 0.25 & $\approx 0$ & Fragmented Baseline \\
Phi-2 & Dec 2023 & 0.20--0.30 & $\approx 0$ & Fragmented Baseline \\
Phi-3-mini & Apr 2024 & 0.90 & $-0.76$ & \textbf{Synthetic Scar} \\
Phi-3.5-mini & Aug 2024 & 0.88 & $-0.74$ & \textbf{Synthetic Scar} \\
Phi-4 & Dec 2024 & 0.30 & $+0.01$ & Adaptive Gearbox \\
\bottomrule
\end{tabular}
\end{table}

The trajectory is striking: \textit{the PTCC is isolated to Phi-3/3.5}. Earlier models (Phi-1.5/2) lack the high-connectivity baseline required for collapse, while the later model (Phi-4) appears to have resolved the vulnerability through a fundamentally different architectural strategy.

\subsection{Phi-1.5 and Phi-2: Pre-Structural Fragmentation}

\begin{figure*}[h!]
    \centering
    \includegraphics[width=\textwidth]{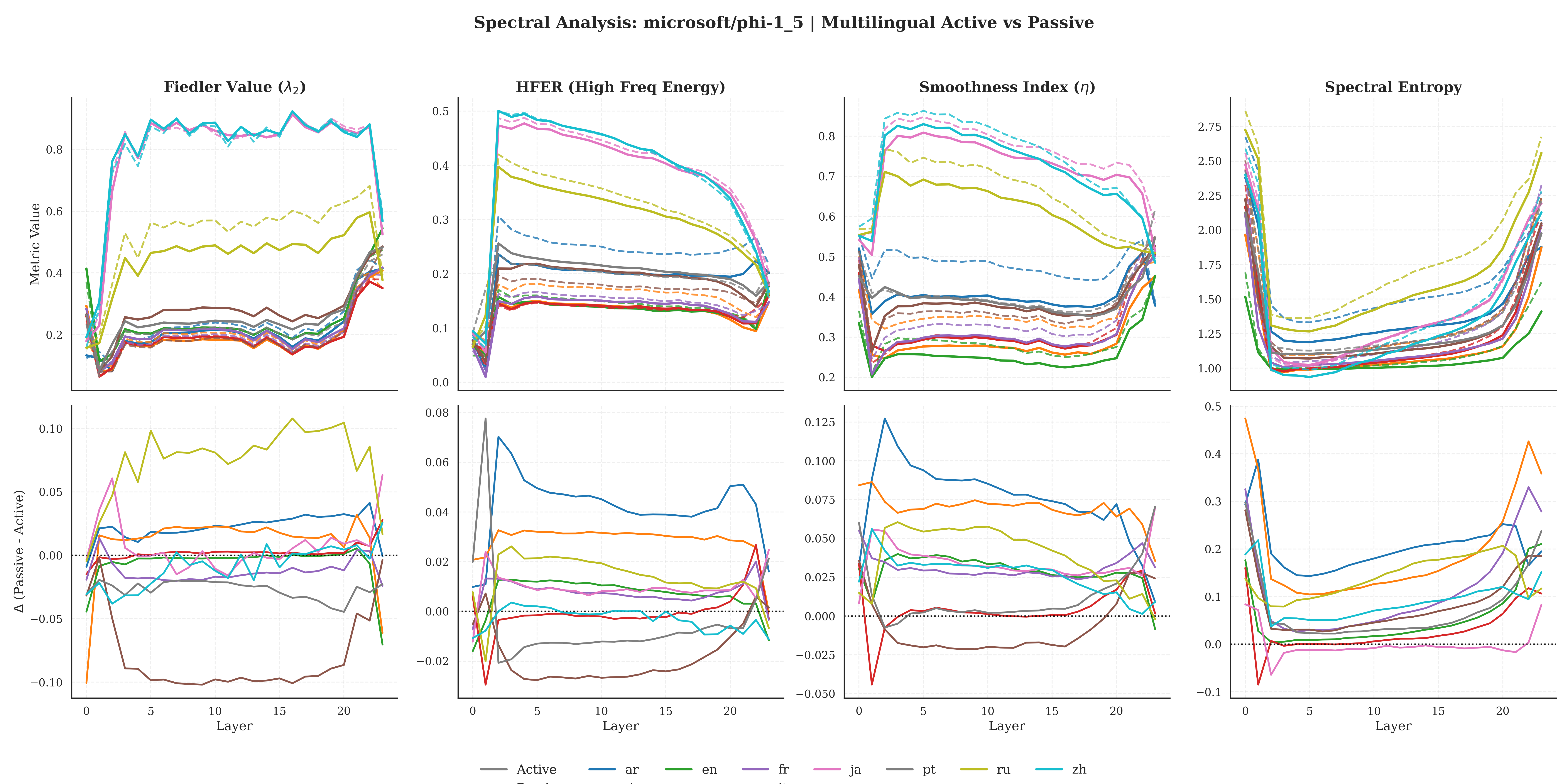}
    \caption{\textbf{Phi-1.5: Pre-Structural Fragmentation.} Unlike Phi-3, English (green) does not collapse under passive voice because it never achieves high connectivity in the first place. Baseline English $\lambda_2 \approx 0.25$ is comparable to passive state, resulting in near-zero delta. Other languages show varied regimes, with some (Russian, Chinese) exhibiting high connectivity. This pattern suggests the model processes English through permanently fragmented pathways rather than specialized circuits that fracture under stress.}
    \label{fig:phi15_spectral}
\end{figure*}

\begin{figure*}[h!]
    \centering
    \includegraphics[width=\textwidth]{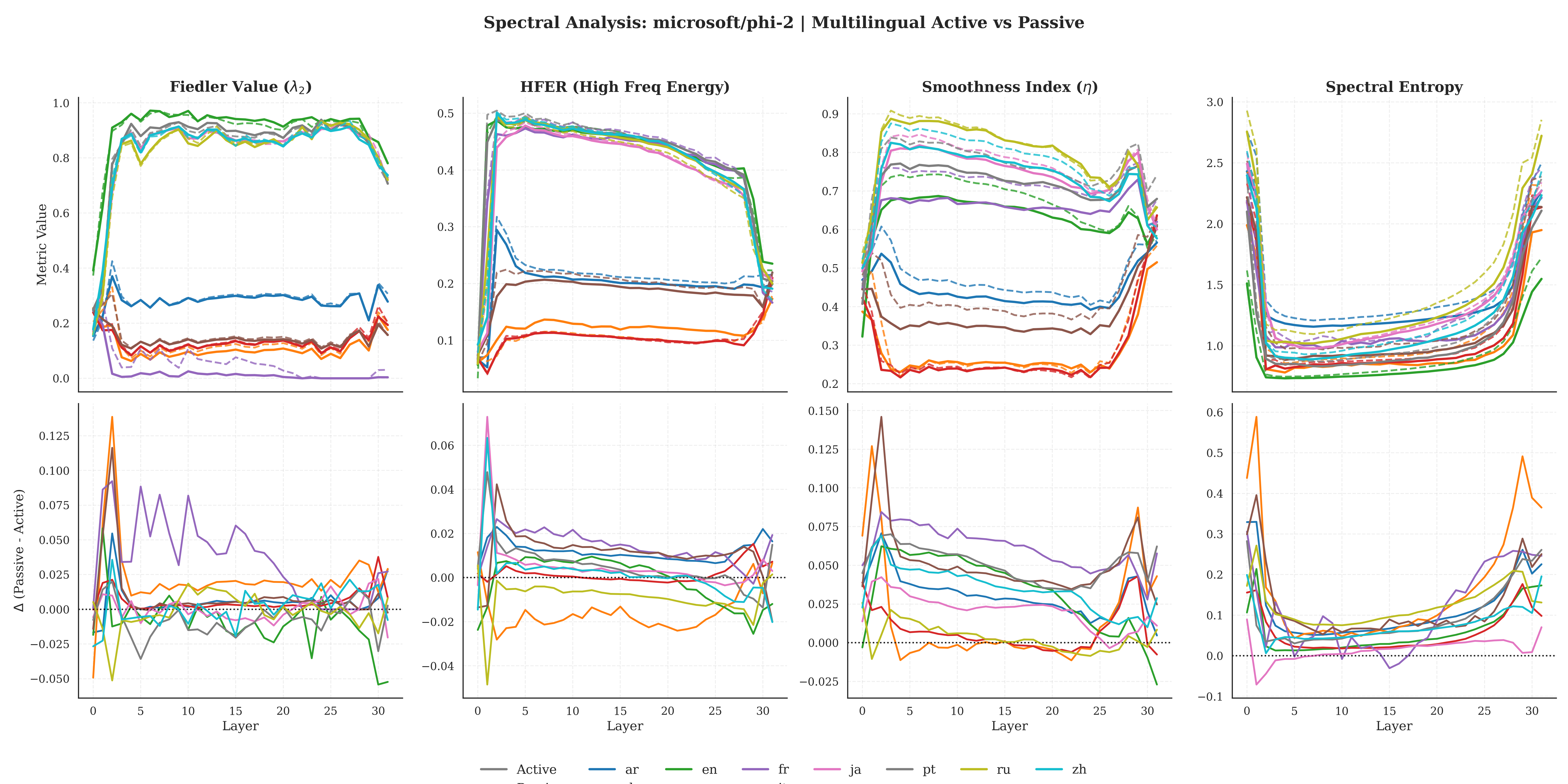}
    \caption{\textbf{Phi-2: Language-Conditioned Regime Separation.} Phi-2 exhibits a striking bimodal distribution: some languages (Russian, Chinese, Japanese) achieve very high connectivity ($\lambda_2 > 0.85$), while English and French remain in a low-connectivity regime ($\lambda_2 \approx 0.20$). Passive voice produces minimal delta because the topology is already language-stratified rather than syntax-sensitive. This represents ``structural apartheid'' rather than syntactic vulnerability.}
    \label{fig:phi2_spectral}
\end{figure*}

Figures~\ref{fig:phi15_spectral} and~\ref{fig:phi2_spectral} show the spectral profiles for Phi-1.5 and Phi-2.

\paragraph{The ``Fragmented Baseline'' Pathology.}
Neither Phi-1.5 nor Phi-2 exhibits the PTCC because they lack the prerequisite: high English connectivity that can collapse. Instead, English is processed through a permanently low-connectivity regime ($\lambda_2 \approx 0.20$--$0.30$), regardless of syntactic construction. We term this the ``Fragmented Baseline'' pathology, the model treats English as a loosely connected collection of tokens rather than an integrated syntactic structure.

This pathology is consistent with the ``Textbooks Are All You Need'' training philosophy \citep{gunasekar2023textbooks}: heavy emphasis on code and formal logic, with English primarily serving as a ``comment language'' rather than a first-class linguistic object. The model learns to process code with high connectivity but relegates English to a simpler, more local processing regime.

\paragraph{Language-Conditioned Regime Separation in Phi-2.}
Phi-2 reveals an additional anomaly: a severe split between languages. Russian, Chinese, and Japanese achieve very high connectivity ($\lambda_2 > 0.85$), while English and French remain fragmented. This raises the question of whether the high-connectivity languages reflect genuine competence or a degenerate ``safe mode'' (uniform attention). We address this via the entropy discriminator in \S\ref{sec:entropy_check}.

\subsection{Phi-4: The Adaptive Gearbox}
\label{sec:phi4}

\begin{figure*}[h!]
    \centering
    \includegraphics[width=\textwidth]{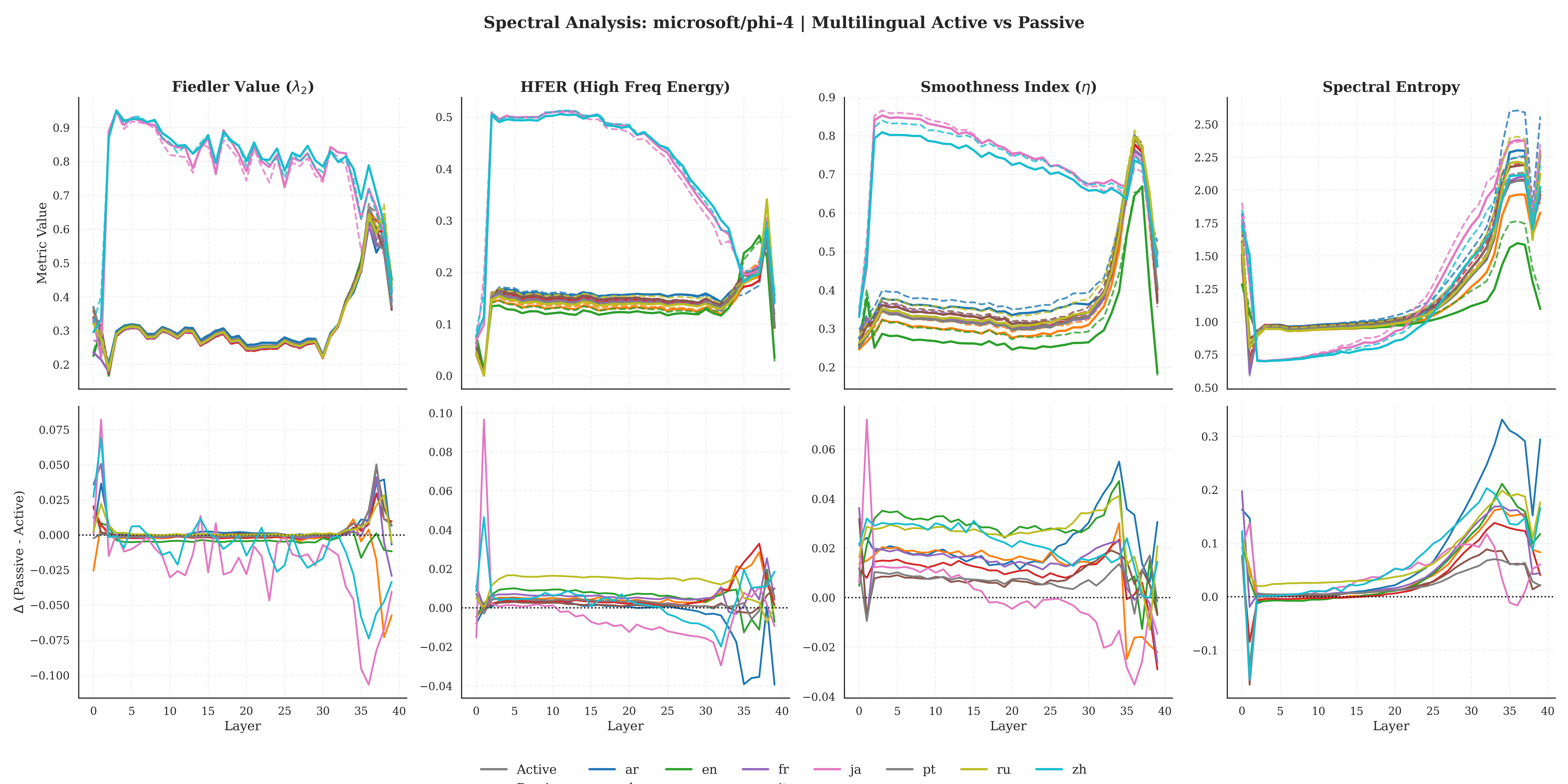}
    \caption{\textbf{Phi-4: The Adaptive Gearbox.} Phi-4 shows no synthetic PTCC ($\Delta\lambda_2 \approx 0$ for passive voice). More strikingly, it exhibits a clean bimodal topology: alphabetic languages (English, French, German) cluster around $\lambda_2 \approx 0.30$, while logographic languages (Japanese, Chinese) cluster around $\lambda_2 \approx 0.93$. Unlike Phi-2's ``panic mode,'' Phi-4's high-connectivity regime is accompanied by \textit{low} entropy, indicating structured integration rather than uniform attention.}
    \label{fig:phi4_spectral}
\end{figure*}

Phi-4 (Figure~\ref{fig:phi4_spectral}) presents a dramatically different profile from both its predecessors and the scarred Phi-3.

\paragraph{Key Observations.}
\begin{enumerate}
    \item \textbf{No PTCC:} Passive voice produces negligible delta ($|\Delta\lambda_2| < 0.02$) across all languages.
    \item \textbf{Bimodal Topology:} Languages split cleanly into two regimes:
    \begin{itemize}
        \item Alphabetic (English, French, German, Russian): $\lambda_2 \approx 0.30$ (``Modular'')
        \item Logographic (Japanese, Chinese): $\lambda_2 \approx 0.93$ (``Dense'')
    \end{itemize}
    \item \textbf{Structured High-Connectivity:} Unlike Phi-2's high-connectivity languages, Phi-4's logographic regime shows \textit{low} spectral entropy (see \S\ref{sec:entropy_check}), indicating intentional global integration rather than degenerate uniformity.
\end{enumerate}

We term this the ``Adaptive Gearbox'', the model switches its internal topology based on the information density of the input script, using modular attention for sparse tokenization (alphabetic) and dense integration for compressed tokenization (logographic).

\subsection{The Entropy Discriminator: Intelligence vs. Panic}
\label{sec:entropy_check}

A high Fiedler value can arise from two mathematically distinct states: (1) \textit{structured integration}, where the model deliberately connects all tokens because each carries high semantic load, or (2) \textit{degenerate uniformity}, where the model connects everything because it lacks priors (``panic mode''). Spectral entropy discriminates between these:

\begin{itemize}
    \item \textbf{Structured Integration:} High $\lambda_2$ + Low Entropy (concentrated eigenvalue mass)
    \item \textbf{Panic Mode:} High $\lambda_2$ + High Entropy (flat eigenvalue spectrum $\approx$ random graph)
\end{itemize}

Table~\ref{tab:entropy_check} presents the entropy analysis.

\begin{table}[h]
\centering
\caption{The Entropy Discriminator. High connectivity in Phi-4 Japanese/Chinese is accompanied by \textit{low} entropy, indicating structured integration. In contrast, Phi-2 Russian shows high entropy despite high connectivity, consistent with panic-mode uniform attention.}
\label{tab:entropy_check}
\begin{tabular}{llccc}
\toprule
Model & Language & Fiedler $\lambda_2$ & Entropy $H_L$ & Interpretation \\
\midrule
Phi-2 & English & 0.94 & 0.77 & Native Competence \\
Phi-2 & Russian & 0.87 & 1.19 & \textbf{Panic Mode} \\
\midrule
Phi-4 & English & 0.28 & 0.96 & Modular Processing \\
Phi-4 & Japanese & 0.84 & 0.86 & \textbf{Structured Integration} \\
Phi-4 & Chinese & 0.84 & 0.84 & \textbf{Structured Integration} \\
\bottomrule
\end{tabular}
\end{table}

The contrast is decisive:
\begin{itemize}
    \item \textbf{Phi-2 Russian:} High $\lambda_2$ (0.87) + High $H_L$ (1.19) = Panic
    \item \textbf{Phi-4 Japanese:} High $\lambda_2$ (0.84) + Low $H_L$ (0.86) = Intelligence
\end{itemize}

Phi-4's entropy for Japanese (0.86) is actually \textit{lower} than its entropy for English (0.96), despite the 3$\times$ higher connectivity. This confirms the high-connectivity regime represents confident, selective attention, not confusion.

\subsection{The Script-Swap Control: Causal Isolation}
\label{sec:script_swap}

To definitively prove that Phi-4's topology adapts to \textit{tokenization density} rather than \textit{semantic content} or \textit{language identity}, we conducted a controlled script-swap experiment. We generated $N=30$ semantically matched triplets, each containing:

\begin{enumerate}
    \item \textbf{Japanese (Kana/Kanji):} Standard logographic representation (dense tokenization)
    \item \textbf{Japanese (Romaji):} Romanized phonetic representation (sparse tokenization)
    \item \textbf{English (Translation):} Semantic equivalent in English (sparse tokenization)
\end{enumerate}

Critically, all sentences include terminal punctuation to control for sentence-boundary recognition effects (pilot experiments revealed that unpunctuated Romaji can be misclassified as code/list content, artificially inflating connectivity).

\begin{table}[h]
\centering
\caption{The Script-Swap Control ($N=30$ triplets). Identical semantic content triggers opposite topological regimes depending solely on script type. Remarkably, Romaji Japanese achieves \textit{identical} connectivity to English, isolating tokenization density as the causal driver.}
\label{tab:script_swap}
\begin{tabular}{lccc}
\toprule
Input Format & Script Type & Fiedler $\lambda_2$ & Regime \\
\midrule
Japanese (Kana/Kanji) & Logographic & 0.78 & Dense Integration \\
Japanese (Romaji) & Latin & 0.25 & Modular Hopping \\
English (Translation) & Latin & 0.25 & Modular Hopping \\
\bottomrule
\end{tabular}
\end{table}

The results are striking: \textbf{same semantics, opposite topology; same script, identical topology}. Switching from Kana to Romaji causes a 3.1$\times$ reduction in connectivity (0.78 $\to$ 0.25). More remarkably, Romaji Japanese achieves \textit{exactly the same} connectivity as English (0.25 vs.\ 0.25), despite encoding different semantic content in a different language. The topological signature is determined entirely by script type.

\begin{figure}[h]
    \centering
    \includegraphics[width=0.9\linewidth]{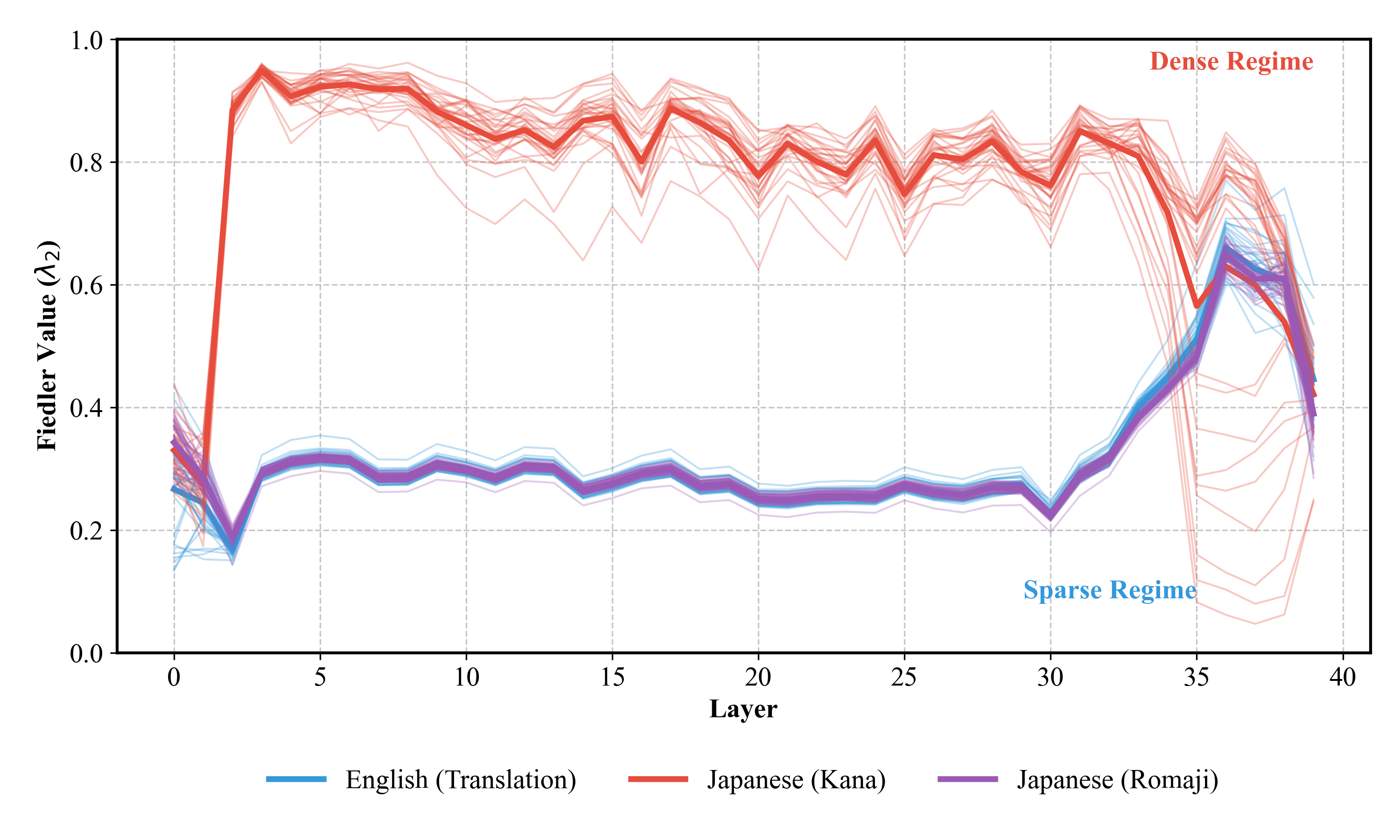}
    \caption{\textbf{The Tokenization Topology Law.} Layer-wise Fiedler trajectories for semantically matched content in three formats. Japanese (Kana, red) maintains high connectivity ($\lambda_2 \approx 0.80$) in the ``Dense Regime,'' while Japanese (Romaji, purple) and English (blue) cluster identically in the ``Sparse Regime'' ($\lambda_2 \approx 0.25$). The phase transition is instantaneous and total: switching script switches topology. Thin lines show individual sentences ($N=30$); bold lines show means. This proves the spectral signature is governed by tokenization mechanics, not semantic content or language identity.}
    \label{fig:tokenization_topology}
\end{figure}

Figure~\ref{fig:tokenization_topology} visualizes this phase transition across all 40 layers. Japanese (Kana) maintains the high-connectivity ``Dense Regime'' ($\lambda_2 \approx 0.80$) throughout mid-layers, while both Romaji and English cluster tightly in the ``Sparse Regime'' ($\lambda_2 \approx 0.25$). The separation is maintained from Layer 3 through Layer 35, with convergence only in the final output layers.

This experiment establishes a causal relationship: \textit{spectral topology tracks information compression, not language identity}. Phi-4 operates as a ``typological gearbox,'' shifting between dense integration (logographic) and modular hopping (alphabetic) based purely on tokenization efficiency. The model has no ``Japanese mode'' or ``English mode'', it has a ``dense token mode'' and a ``sparse token mode,'' selected automatically based on the physical properties of the input stream.

\subsection{The Tokenization Topology Law}

Synthesizing the Phi lineage analysis and the script-swap control, we propose the following empirical regularity:

\begin{quote}
\textbf{Tokenization Topology Law (Hypothesis):} Attention graph connectivity scales inversely with token count for equivalent semantic content. Dense tokenization (logographic scripts) triggers global integration; sparse tokenization (alphabetic scripts) triggers modular bridging.
\end{quote}

The script-swap experiment provides unusually clean causal evidence for this law. Three predictions follow directly:

\begin{enumerate}
    \item \textit{Logographic scripts should show higher connectivity than alphabetic scripts.} Confirmed: Japanese (Kana) $\lambda_2 = 0.78$ vs.\ English $\lambda_2 = 0.25$.
    
    \item \textit{Romanization should collapse logographic connectivity to alphabetic levels.} Confirmed: Romaji $\lambda_2 = 0.25$, identical to English.
    
    \item \textit{The regime should track script, not language.} Confirmed: Romaji Japanese is topologically indistinguishable from English despite encoding different semantic content.
\end{enumerate}

The law remains a hypothesis pending validation across additional models and scripts (e.g., Arabic, Hebrew, Thai). However, the identity $\lambda_2^{\text{Romaji}} = \lambda_2^{\text{English}} = 0.25$ provides striking evidence that topology adapts to information density rather than reflecting fixed language-specific circuitry. The model appears to operate a content-agnostic ``compression detector'' that routes dense token streams through global integration and sparse streams through modular bridging.

\subsection{Evolutionary Interpretation: From Trauma to Maturity}

The complete Phi lineage can now be understood as an evolutionary trajectory through distinct spectral pathologies:

\begin{enumerate}
    \item \textbf{Phi-1.5/2 (Infancy):} ``Structural Apartheid.'' English is undertrained, processed through permanently fragmented pathways. High connectivity on some languages reflects panic (high entropy), not competence.
    
    \item \textbf{Phi-3/3.5 (Adolescence):} ``The Synthetic Scar.'' Microsoft attempted to bridge the fragmented English baseline with new multilingual capacity. The resulting architecture achieved high English connectivity but compressed routing into sparse compensatory circuits that fracture under syntactic stress.
    
    \item \textbf{Phi-4 (Maturity):} ``The Adaptive Gearbox.'' The PTCC is healed. The model operates a content-agnostic topology selector: dense token streams (logographic) receive global integration ($\lambda_2 \approx 0.78$), sparse streams (alphabetic) receive modular bridging ($\lambda_2 \approx 0.25$). The script-swap experiment proves this is a physical response to tokenization density, not language-specific circuitry.
\end{enumerate}

The PTCC, in this view, was a developmental injury, the price of attempting efficiency gains during a curriculum transition. Phi-4's resolution suggests Microsoft identified and corrected the underlying failure mode, replacing brittle language-specific circuits with a robust density-adaptive mechanism.

\subsection{Implications for the Main Results}

The Phi lineage analysis has several implications for interpreting our main findings:

\paragraph{1. The PTCC is curriculum-specific.}
The synthetic PTCC should not be interpreted as a universal signature of synthetic training. It appears to emerge from specific curriculum transitions that compress routing redundancy into sparse circuits. Other synthetic training approaches (e.g., Phi-1.5/2's code-focused curriculum) produce different pathologies.

\paragraph{2. Spectral forensics reveals developmental history.}
The progression Fragmented $\to$ Scarred $\to$ Adaptive demonstrates that spectral analysis can recover not just training regime (synthetic vs. organic) but developmental trajectory within a model family. This opens possibilities for ``spectral archaeology'', reconstructing training history from weight inspection alone.

\paragraph{3. The typological gearbox suggests deeper structure.}
Phi-4's script-dependent topology suggests that advanced models may encode \textit{typological awareness}, implicit knowledge of how different writing systems encode information. This raises questions about whether such awareness is learned, emergent, or architecturally imposed.

\section{Spectral Profiles Across Model Families}
\label{app:spectral_profiles}

This appendix presents the complete spectral signatures for all eight primary models evaluated in our study. These visualizations enable direct comparison of processing strategies across architectural families and provide the empirical basis for the forensic classification rules presented in the main text.

%%%%%%%%%%%%%%%%%%%%%%%%%%%%%%%%%%%%%%%%%%%%%%%%%%%%%%%%%%%%%%%%%%%%%%%%%%%%%%%
% FAMILY 1: THE SYNTHETIC SCAR (PHI)
%%%%%%%%%%%%%%%%%%%%%%%%%%%%%%%%%%%%%%%%%%%%%%%%%%%%%%%%%%%%%%%%%%%%%%%%%%%%%%%

\subsection{Strategy A: English-Specialized Collapse (Phi-3)}

The Phi-3 family exemplifies the spectral signature of the synthetic PTCC. These models achieve strong benchmark performance through efficient, specialized circuits, but this efficiency comes at the cost of robustness to syntactic variation.

\begin{figure*}[h!]
    \centering
    \includegraphics[width=\textwidth]{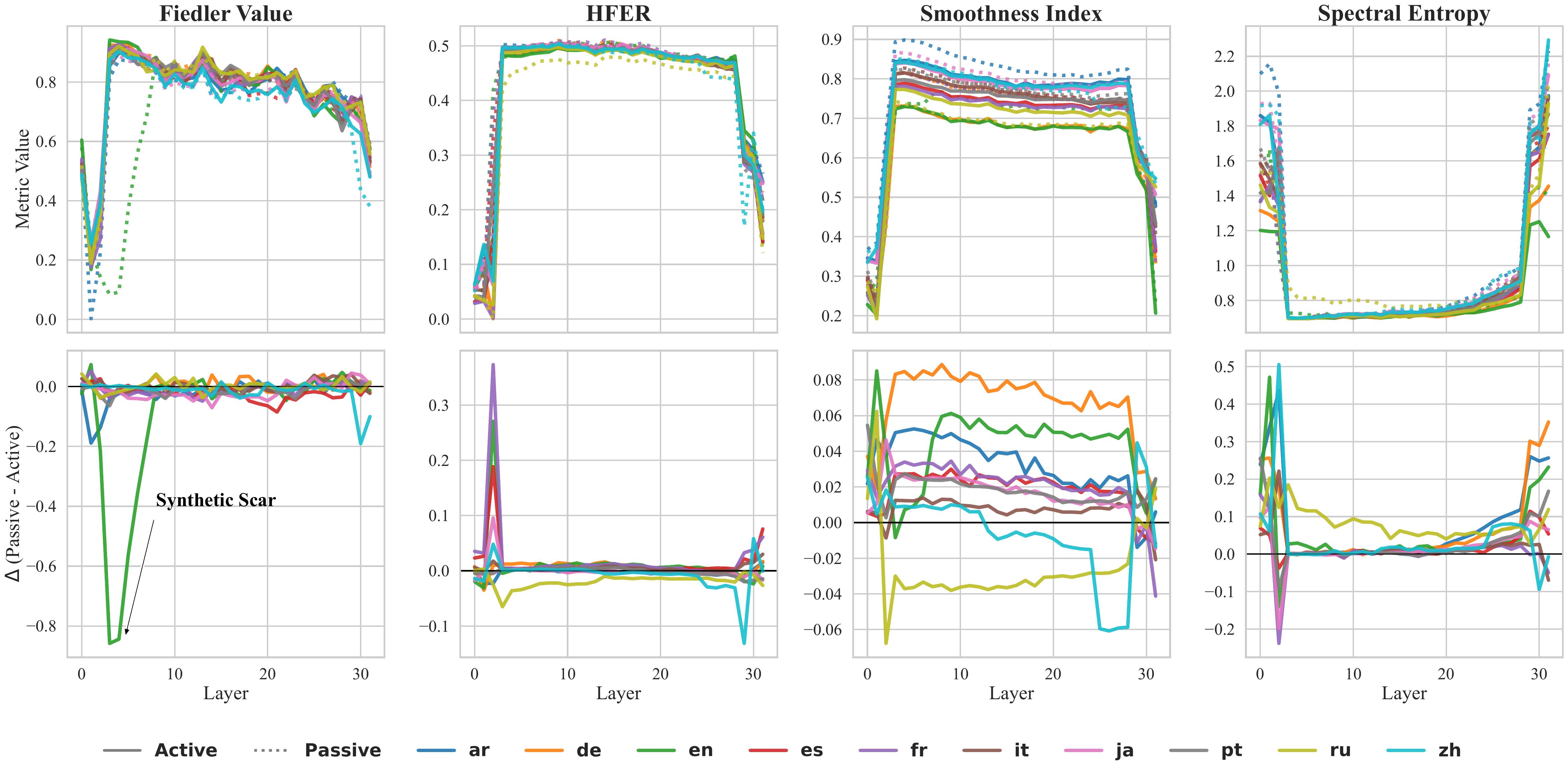}
    \caption{\textbf{The Synthetic Scar (Phi-3-mini).} Top row: raw spectral metrics across 32 layers. Bottom row: passive minus active deltas. English (green) exhibits catastrophic Fiedler collapse at layers 2--4 ($\lambda_2: 0.90 \to 0.14$, minimum 0.08 at Layer 3), while all other languages remain within $|\Delta\lambda_2| < 0.06$. The concurrent HFER spike and entropy increase confirm topological fragmentation accompanies the connectivity loss. This English-specific signature enables forensic identification of the scarred training regime.}
    \label{fig:phi3_spectral}
\end{figure*}

\paragraph{The Collapse Mechanism.}
Figure~\ref{fig:phi3_spectral} reveals the characteristic ``synthetic scar'' in Phi-3-mini. Under active voice, English maintains healthy connectivity ($\lambda_2 \approx 0.90$) comparable to other languages. Under passive voice stress, English connectivity collapses catastrophically: $\lambda_2$ drops to 0.14 at Layer 2 and reaches a minimum of 0.08 at Layer 3, a 91\% reduction. The collapse spans layers 1--5 before recovering by Layer 6. Critically, all nine other languages remain stable ($|\Delta\lambda_2| < 0.06$), demonstrating that this is an English-specific vulnerability rather than a general syntactic limitation.

The concurrent HFER spike (second column) confirms that the collapse involves signal fragmentation into high-frequency modes, while the entropy increase (fourth column) indicates loss of structured computation. Together, these metrics paint a picture of attention graphs fracturing into disconnected local clusters unable to integrate global context.

\begin{figure*}[h!]
    \centering
    \includegraphics[width=\textwidth]{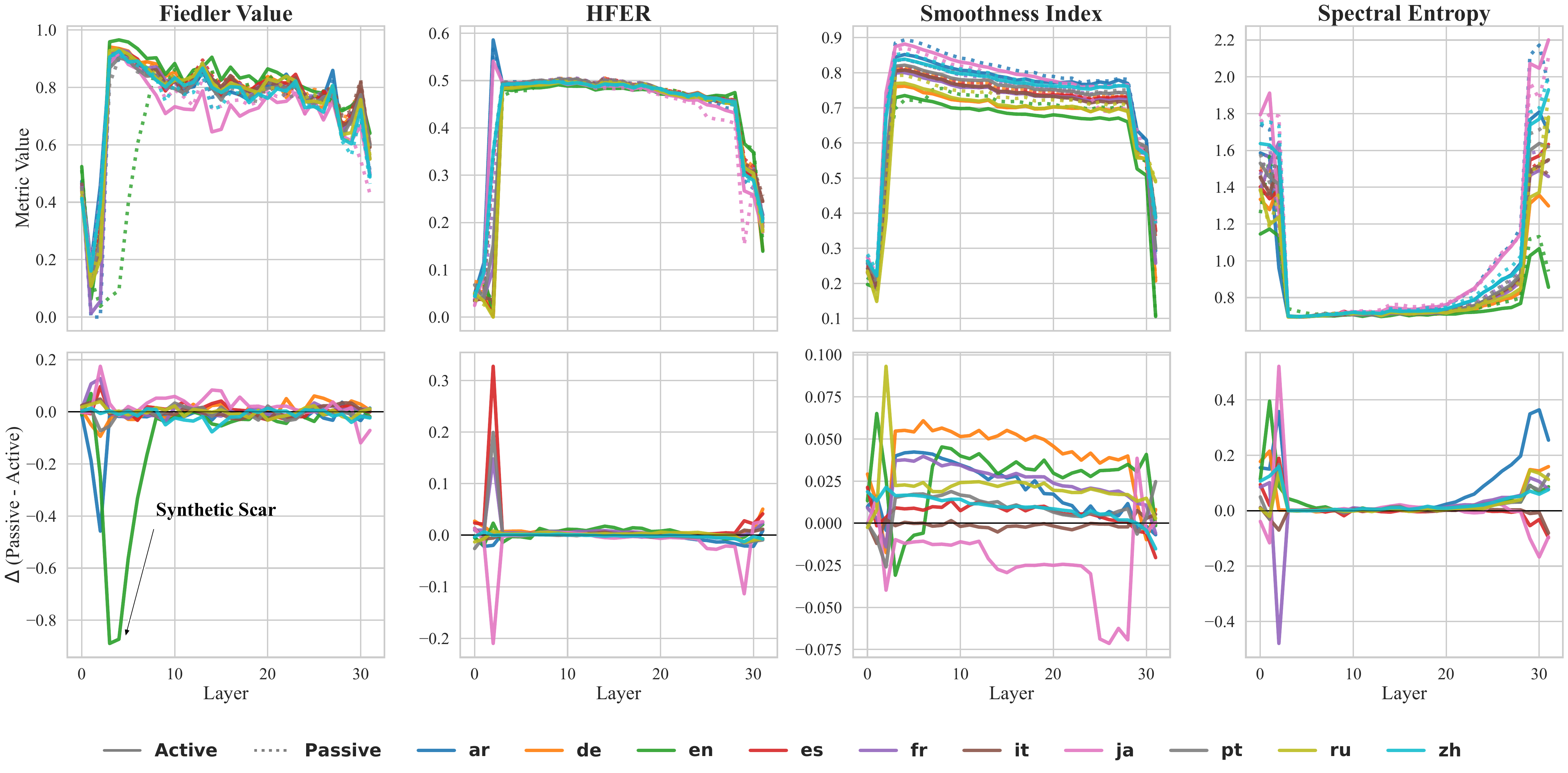}
    \caption{\textbf{Replication of the Synthetic Scar (Phi-3.5-mini).} The collapse pattern replicates precisely in Phi-3.5 ($\overline{\Delta\lambda_2}_{[2,5]} = -0.64$), with identical layer localization (2--4), language specificity (English only), and recovery dynamics. This cross-version consistency confirms the PTCC is a family-level characteristic driven by the synthetic training curriculum, not a version-specific anomaly.}
    \label{fig:phi35_spectral}
\end{figure*}

\paragraph{Replication Across Versions.}
Figure~\ref{fig:phi35_spectral} demonstrates that the synthetic PTCC is a persistent family characteristic. Phi-3.5-mini exhibits an identical collapse pattern ($\overline{\Delta\lambda_2}_{[2,5]} = -0.64$), with the same layer localization and language specificity. This replication across model versions rules out version-specific artifacts and confirms that the PTCC originates from the training data curriculum rather than architectural idiosyncrasies.

\newpage

%%%%%%%%%%%%%%%%%%%%%%%%%%%%%%%%%%%%%%%%%%%%%%%%%%%%%%%%%%%%%%%%%%%%%%%%%%%%%%%
% FAMILY 2: ORGANIC ROBUSTNESS (LLAMA & MISTRAL)
%%%%%%%%%%%%%%%%%%%%%%%%%%%%%%%%%%%%%%%%%%%%%%%%%%%%%%%%%%%%%%%%%%%%%%%%%%%%%%%

\subsection{Strategy C: Cross-Linguistic Uniformity (Llama)}

The Llama family demonstrates the spectral signature of broad-spectrum organic pre-training: uniform processing pathways that remain stable across languages and syntactic transformations.

\begin{figure*}[h!]
    \centering
    \includegraphics[width=\textwidth]{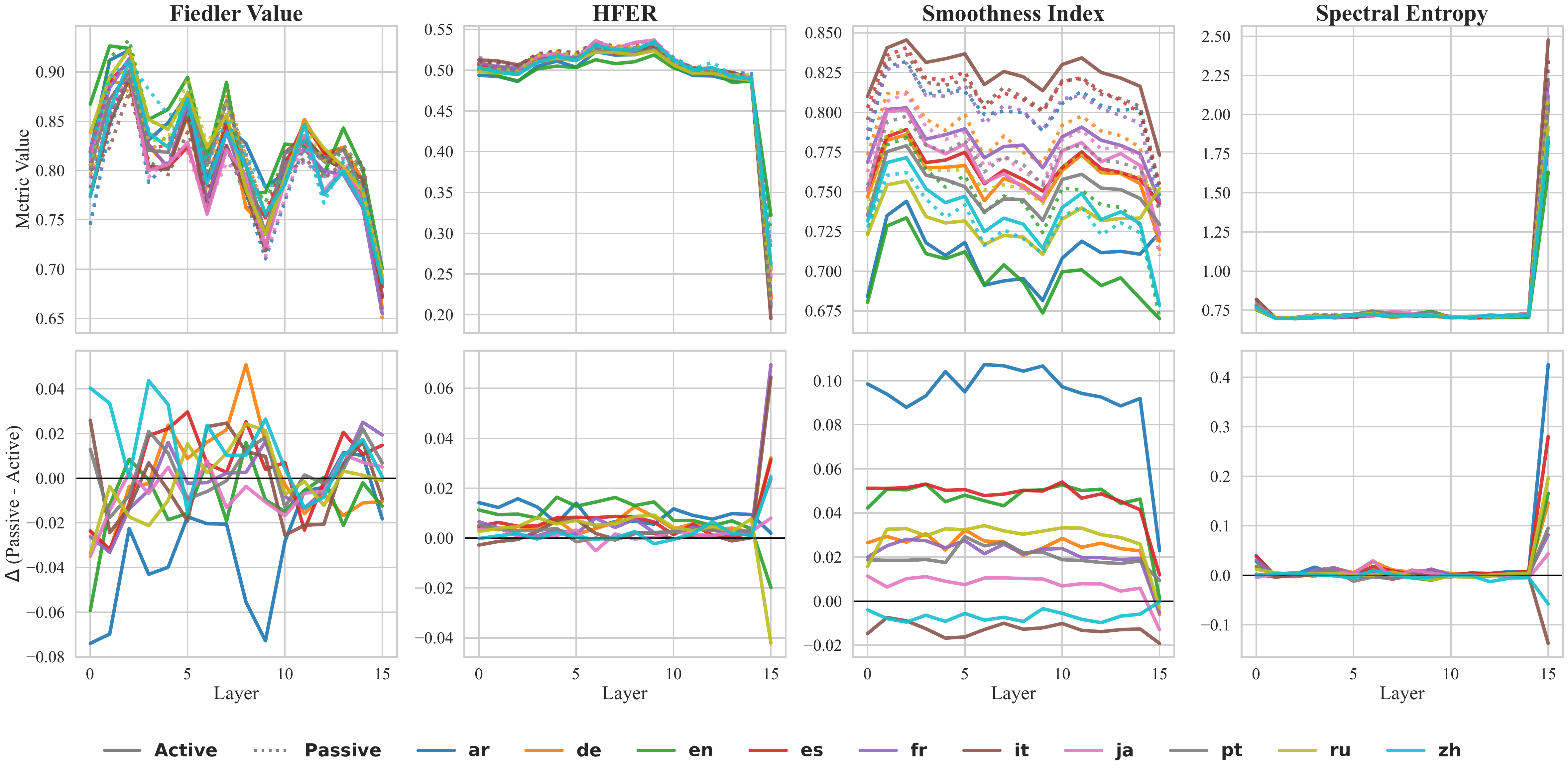}
    \caption{\textbf{Cross-Linguistic Uniformity (Llama-3.2-1B).} All ten languages maintain stable, high connectivity ($\lambda_2 \approx 0.80$) with minimal passive-active deltas ($|\Delta\lambda_2| < 0.02$). No language-specific fragility emerges. This uniform strategy reflects broad-spectrum organic training that builds redundant pathways capable of handling syntactic variation without topological disruption.}
    \label{fig:llama1b_spectral}
\end{figure*}

\paragraph{The Uniform Baseline.}
Figure~\ref{fig:llama1b_spectral} shows Llama-3.2-1B maintaining high algebraic connectivity ($\lambda_2 \approx 0.80$) across all ten languages under both active and passive voice. The delta panel (bottom row) reveals near-zero shifts for all language-construction pairs, with no language exhibiting preferential vulnerability. This uniformity suggests that exposure to diverse, organic web data forces the model to learn attention mechanisms invariant to local syntactic perturbations, robustness through redundancy rather than specialization.

\begin{figure*}[h]
    \centering
    \includegraphics[width=\textwidth]{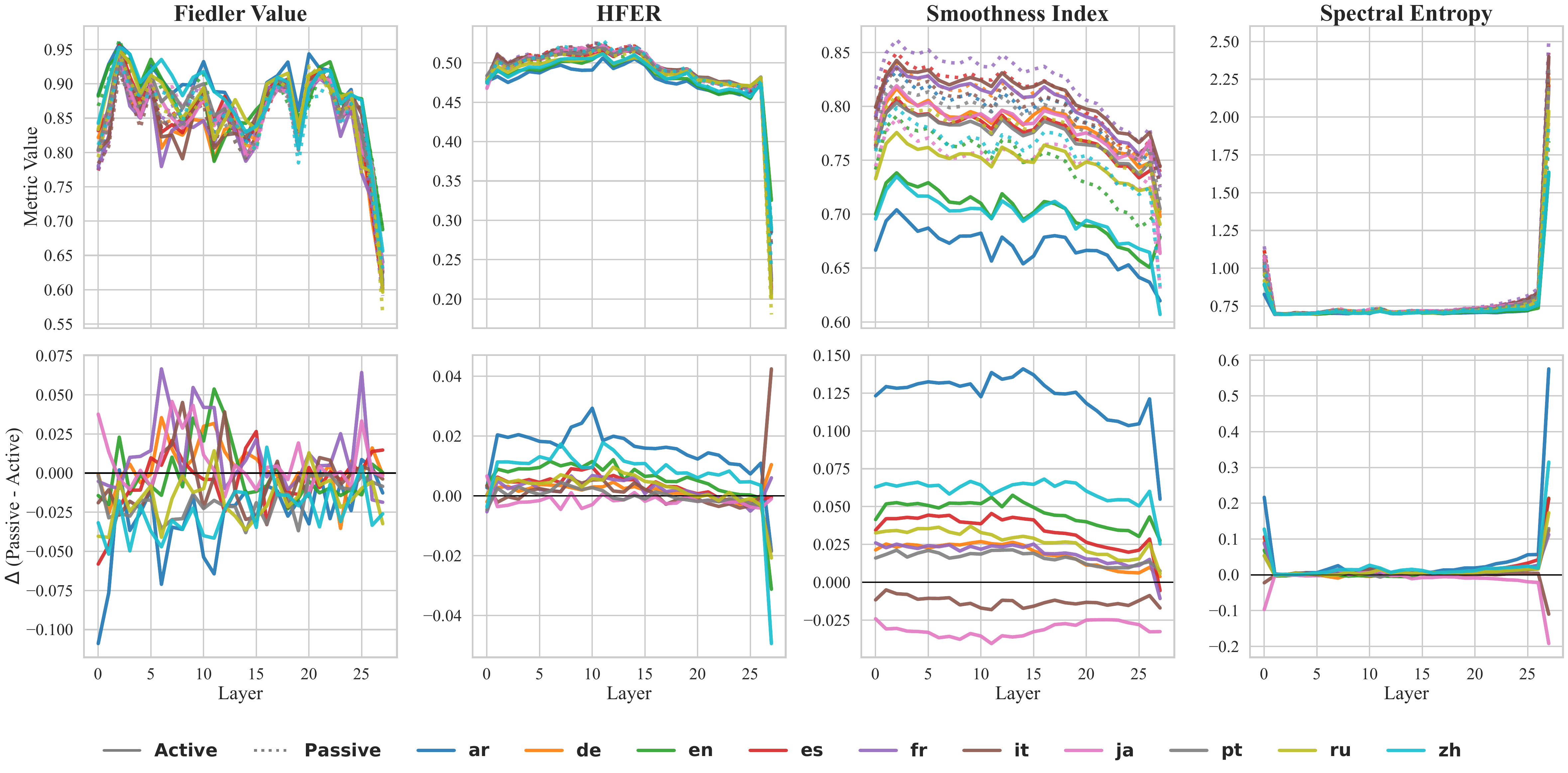}
    \caption{\textbf{Scale Invariance (Llama-3.2-3B).} The uniform strategy persists at 3B parameters. Connectivity remains high ($\lambda_2 \approx 0.80$) and language-agnostic, with smoother layer-wise trajectories than the 1B model. Scaling capacity does not introduce specialization scars; the organic training distribution produces consistent robustness across scales.}
    \label{fig:llama3b_spectral}
\end{figure*}

\paragraph{Scale Invariance.}
Figure~\ref{fig:llama3b_spectral} confirms that the uniform strategy persists at larger scale. Llama-3.2-3B exhibits the same high-connectivity baseline and minimal language-specific effects as its 1B counterpart. Increased capacity does not induce specialization or create new vulnerabilities, the model simply maintains its robust topology with slightly smoother spectral trajectories. This scale invariance suggests the uniform strategy is an emergent property of the training distribution rather than a capacity-limited compromise.

\subsection{Strategy D: Confidence Calibration (Mistral)}

Mistral-7B presents a distinct variant of organic robustness characterized by an unusual relationship between syntactic stress and uncertainty.

\begin{figure*}[h!]
    \centering
    \includegraphics[width=\textwidth]{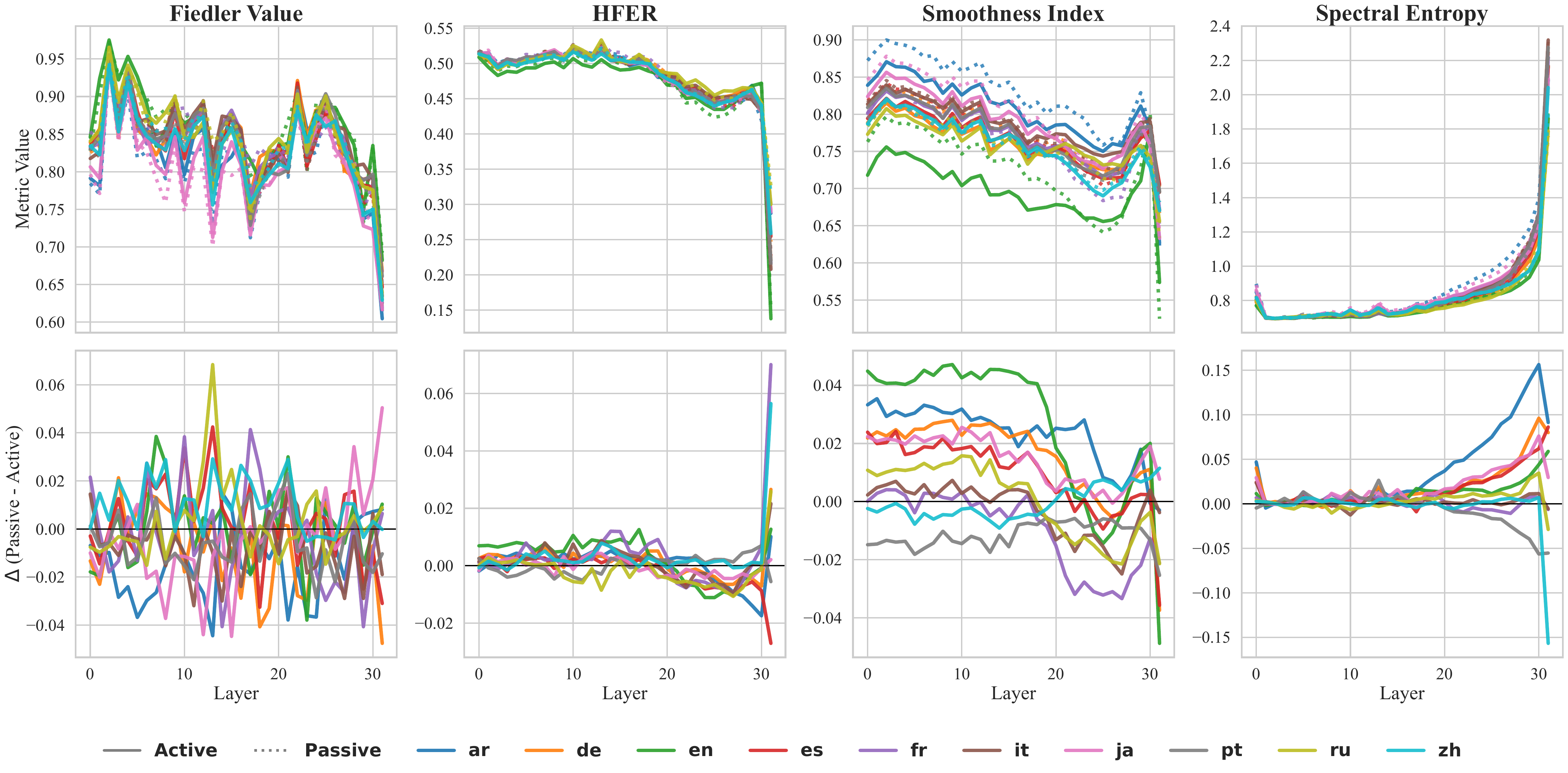}
    \caption{\textbf{Confidence Inversion (Mistral-7B).} Mistral maintains high connectivity ($\lambda_2 \approx 0.85$) with minimal Fiedler or HFER shifts under passive voice. Uniquely, spectral entropy \textit{decreases} ($\Delta\text{Ent} \approx -0.15$), suggesting passive syntax reduces rather than increases uncertainty. This ``confident'' strategy treats syntactic constraints as disambiguation cues, producing a distinct robustness profile from Llama's uniform approach.}
    \label{fig:mistral_spectral}
\end{figure*}

\paragraph{The Confidence Inversion.}
Figure~\ref{fig:mistral_spectral} reveals that Mistral maintains stable Fiedler values ($\lambda_2 \approx 0.85$) and HFER across syntactic transformations, but exhibits a unique pattern in spectral entropy (fourth column). While most models show increased entropy (higher uncertainty) under passive voice, Mistral shows a \textit{decrease} ($\Delta\text{Ent} \approx -0.15$). 

We interpret this as a ``confidence inversion'': the rigid constraints of passive syntax (mandatory auxiliary verbs, fixed word order) provide additional structural anchors that \textit{reduce} rather than increase parsing ambiguity. Rather than fracturing the attention graph, passive constructions crystallize it into a more deterministic configuration. This behavior likely reflects Mistral's architectural choices (Sliding Window Attention, Group Query Attention) interacting with high-quality training data to produce a qualitatively different form of robustness.

%%%%%%%%%%%%%%%%%%%%%%%%%%%%%%%%%%%%%%%%%%%%%%%%%%%%%%%%%%%%%%%%%%%%%%%%%%%%%%%
% FAMILY 3: MULTILINGUAL FRAGMENTATION (QWEN)
%%%%%%%%%%%%%%%%%%%%%%%%%%%%%%%%%%%%%%%%%%%%%%%%%%%%%%%%%%%%%%%%%%%%%%%%%%%%%%%

\subsection{Strategy B: English-Fragmented Baseline (Qwen)}

The Qwen family exhibits a third distinct strategy: permanently fragmented English processing that differs qualitatively from both the Phi collapse and the Llama/Mistral robustness.

\begin{figure*}[h!]
    \centering
    \includegraphics[width=\textwidth]{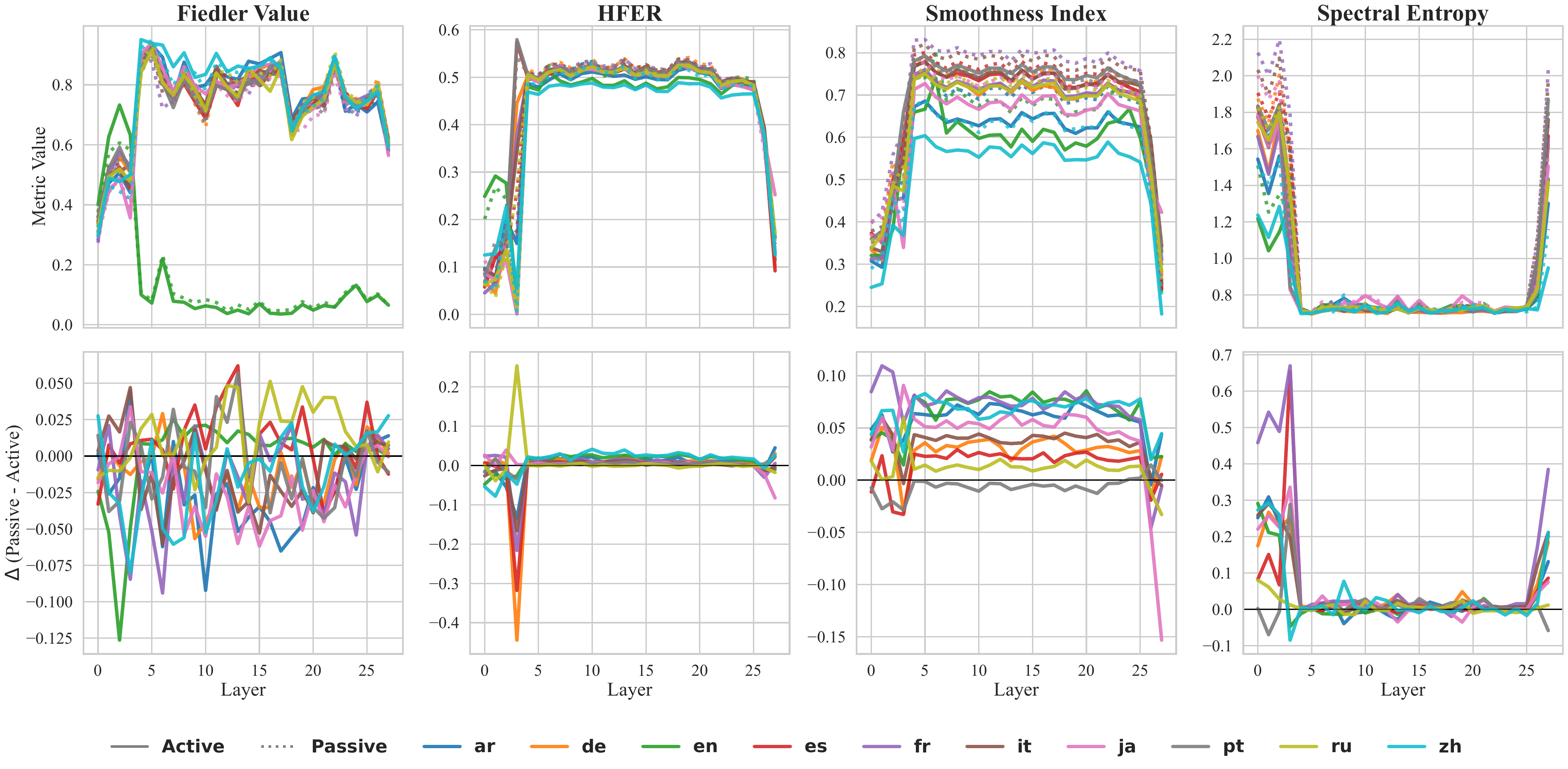}
    \caption{\textbf{English-Fragmented Baseline (Qwen-2.5-7B).} English (green) exhibits permanently low connectivity ($\lambda_2 \approx 0.12$) even under active voice, while other languages maintain $\lambda_2 \approx 0.70$. This is not stress-induced collapse but a baseline architectural separation. The inverted HFER response ($\Delta\text{HFER} < 0$) confirms English is processed via a qualitatively different pathway. Unlike the Phi PTCC, this represents stable fragmentation rather than brittle specialization.}
    \label{fig:qwen7b_spectral}
\end{figure*}

\paragraph{The Fragmented Pathway.}
Figure~\ref{fig:qwen7b_spectral} shows that Qwen-2.5-7B processes English through a fundamentally different topological regime. Even under \textit{active} voice, English connectivity is severely reduced ($\lambda_2 \approx 0.12$) compared to other languages ($\lambda_2 \approx 0.70$). This is not a stress-induced collapse but a permanent baseline difference, English occupies a distinct, low-connectivity subspace of the attention manifold.

The HFER response is also inverted: passive voice \textit{reduces} high-frequency energy ($\Delta\text{HFER} \approx -0.35$) rather than increasing it. This suggests Qwen has learned a specialized English processing pathway that operates via different computational mechanisms than its multilingual circuits. Whether this reflects deliberate architectural choices, tokenizer effects, or training data composition remains an open question.

\begin{figure*}[h!]
    \centering
    \includegraphics[width=\textwidth]{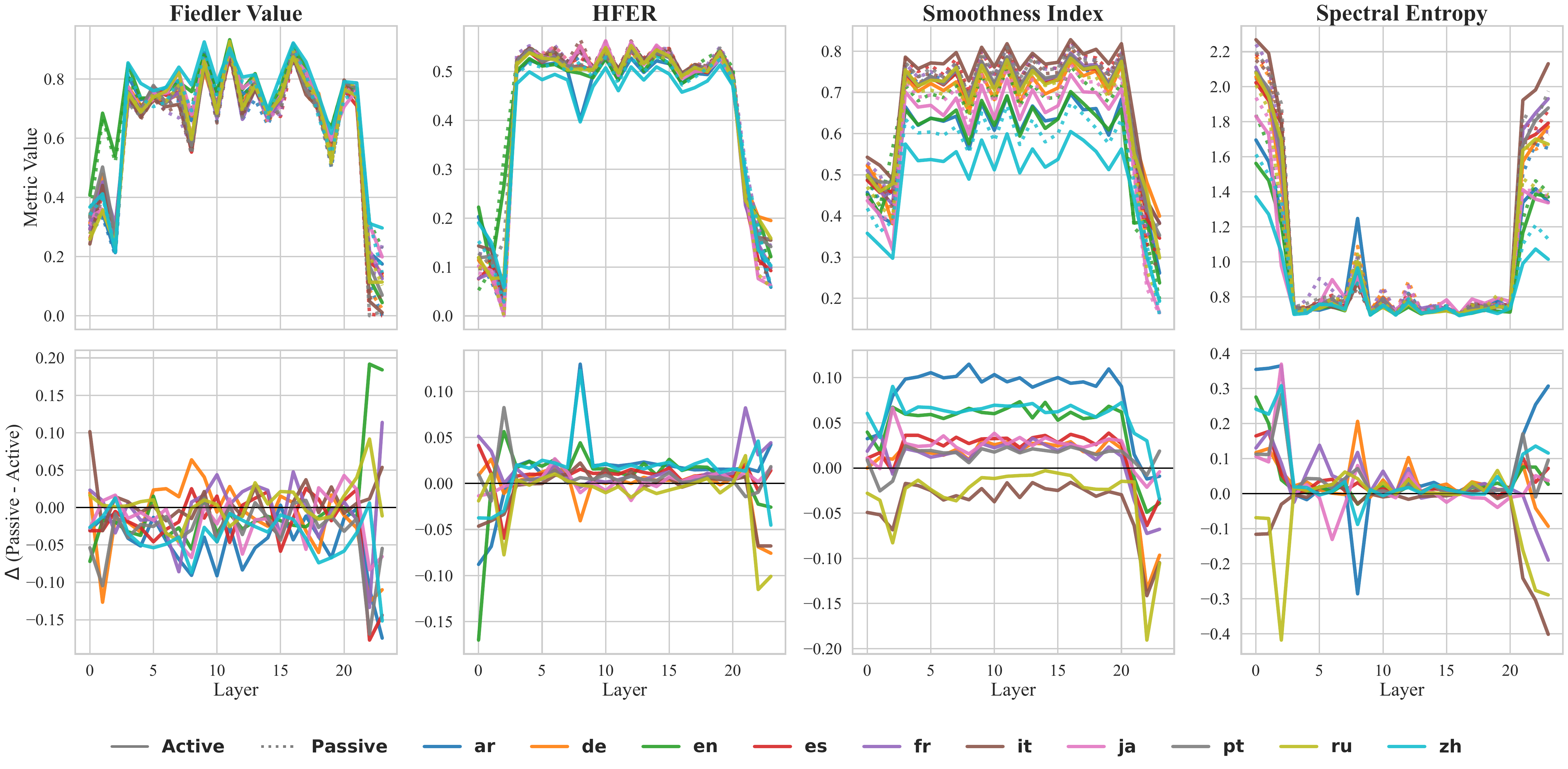}
    \caption{\textbf{Scale Consistency (Qwen-2.5-0.5B).} The fragmented English baseline persists at 0.5B parameters, confirming this is a family-level characteristic independent of model capacity. English connectivity remains low ($\lambda_2 \approx 0.15$) relative to other languages across all layers. This consistency across scales suggests the fragmentation originates from tokenizer or data mixture choices rather than capacity constraints.}
    \label{fig:qwen05b_spectral}
\end{figure*}

\paragraph{Scale Consistency.}
Figure~\ref{fig:qwen05b_spectral} demonstrates that the fragmented English baseline persists down to the smallest Qwen variant. At 0.5B parameters, the same pattern emerges: low English connectivity at baseline, minimal stress-induced change, inverted HFER response. This scale consistency indicates the fragmentation is a fundamental property of the Qwen family, likely originating from tokenizer design or pre-training mixture, rather than an emergent property of scale.

\begin{figure*}[h!]
    \centering
    \includegraphics[width=\textwidth]{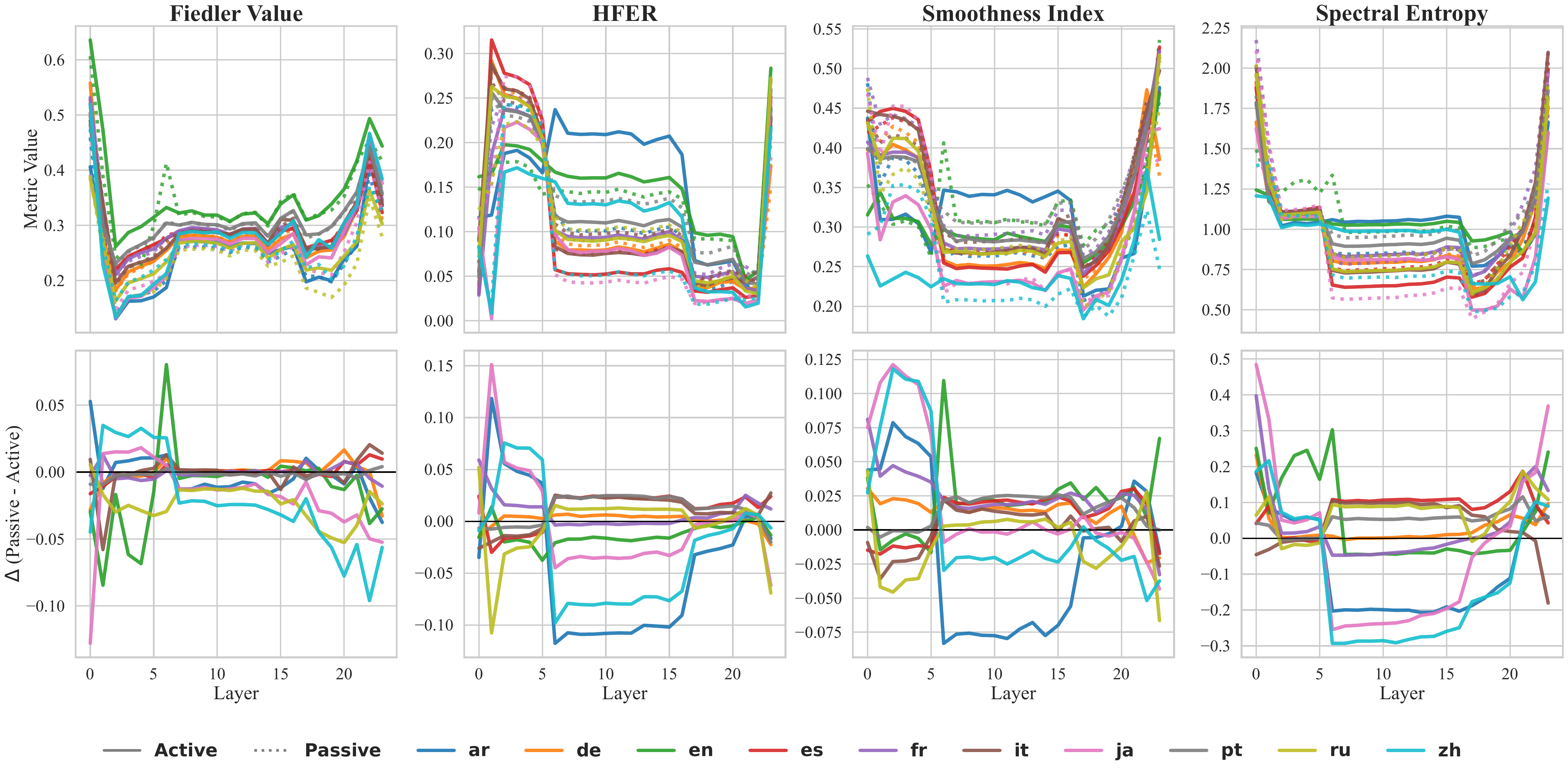}
    \caption{\textbf{Mixture-of-Experts Dynamics (Qwen-1.5-MoE).} The MoE architecture introduces higher spectral variance (jagged trajectories) due to discrete router decisions, but preserves the Qwen family's fragmented English strategy. No synthetic PTCC emerges despite the efficiency-focused sparse architecture, confirming the PTCC is a property of synthetic \textit{training data} rather than parameter-efficient \textit{architectures}.}
    \label{fig:qwenmoe_spectral}
\end{figure*}

\paragraph{Mixture-of-Experts Dynamics.}
Figure~\ref{fig:qwenmoe_spectral} extends the analysis to Qwen-1.5-MoE. The MoE architecture introduces higher variance in spectral trajectories (visible as jagged layer-wise profiles) due to discrete expert routing decisions. However, the fundamental strategy aligns with dense Qwen models: fragmented English baseline, inverted HFER response, no catastrophic collapse. Notably, the MoE shows no evidence of the synthetic PTCC despite its efficiency-focused architecture, suggesting the PTCC is specific to synthetic \textit{data} rather than efficient \textit{architectures}.

%%%%%%%%%%%%%%%%%%%%%%%%%%%%%%%%%%%%%%%%%%%%%%%%%%%%%%%%%%%%%%%%%%%%%%%%%%%%%%%
% SUMMARY
%%%%%%%%%%%%%%%%%%%%%%%%%%%%%%%%%%%%%%%%%%%%%%%%%%%%%%%%%%%%%%%%%%%%%%%%%%%%%%%

\subsection{Summary: Four Strategies, One Framework}

The spectral profiles presented above demonstrate that a single analytical framework, attention-graph connectivity under syntactic stress, reveals four qualitatively distinct processing strategies:

\begin{enumerate}
    \item \textbf{Specialized Collapse (Phi-3):} High baseline connectivity with catastrophic English-specific failure under stress. Signature of curriculum-induced brittle, low-redundancy circuits.
    
    \item \textbf{Fragmented Baseline (Qwen):} Permanently low English connectivity at baseline, stable under stress. Signature of architectural or data-driven pathway separation.
    
    \item \textbf{Cross-Linguistic Uniformity (Llama):} High connectivity maintained uniformly across languages and constructions. Signature of organic training building redundant, stress-resistant pathways.
    
    \item \textbf{Confidence Calibration (Mistral):} High connectivity with entropy reduction under stress. Signature of architectures that leverage syntactic constraints for disambiguation.
\end{enumerate}

These strategies are invisible to behavioral benchmarks but carry direct implications for deployment: multilingual applications should prefer uniform-strategy models, while applications requiring syntactic flexibility should avoid specialized-collapse models regardless of their headline performance metrics.

\section{Experimental Setup}
\label{sec:experimental_setup}

\subsection{Models}
\label{subsec:models}
We conduct experiments across multiple language model families and architectures (dense and sparse) to ensure the generalizability of our findings. All models are loaded in FP16 precision using the Hugging Face \texttt{transformers} library.

\paragraph{Phi Family.} We evaluate the evolution of the Phi architecture:
\begin{itemize}[nosep]
    \item \textbf{Phi-3:} \texttt{microsoft/Phi-3-mini-4k-instruct} (3.8B, 32 layers).
    \item \textbf{Phi-3.5:} \texttt{microsoft/Phi-3.5-mini-instruct} (3.8B), evaluating improvements in long-context and reasoning capabilities.
\end{itemize}

\paragraph{Llama Family.} We evaluate both base and instruction-tuned variants at two scales:
\begin{itemize}[nosep]
    \item \textbf{1B Scale:} \texttt{meta-llama/Llama-3.2-1B} and \texttt{Llama-3.2-1B-Instruct}.
    \item \textbf{3B Scale:} \texttt{meta-llama/Llama-3.2-3B} and \texttt{Llama-3.2-3B-Instruct}.
\end{itemize}

\paragraph{Qwen Family.} To test robustness across varying parameter counts and multilingual capabilities, we encompass:
\begin{itemize}[nosep]
    \item \textbf{Qwen 2.5:} \texttt{Qwen/Qwen2.5-0.5B-Instruct} and \texttt{Qwen/Qwen2.5-7B-Instruct}.
\end{itemize}

\paragraph{Mistral Family.} We evaluate \texttt{mistralai/Mistral-7B-Instruct-v0.1} (7B) to benchmark spectral properties against another leading dense architecture.

\paragraph{Phi Family (Extended).} Beyond Phi-3, we analyze the architectural evolution:
\begin{itemize}[nosep]
    \item \textbf{Phi-4:} \texttt{microsoft/Phi-4} (14B), analyzing topological changes in newer dense models.
\end{itemize}

\paragraph{Evolution Analysis.} To study architectural evolution, we additionally include:
\begin{itemize}[nosep]
    \item \texttt{openlm-research/open\_llama\_3b} (Llama-1 proxy, trained on RedPajama)
    \item \texttt{openlm-research/open\_llama\_3b\_v2} (Llama-2 architecture proxy)
\end{itemize}
\subsection{Datasets}
\label{subsec:datasets}
\subsubsection{Syntactic Integrity Benchmark}
We construct controlled sentence pairs to isolate syntactic structure effects on attention topology:
\paragraph{Voice Contrast.} 200 active and 200 passive voice sentence pairs matched for semantic content:
\begin{quote}
\small
\textit{Active:} ``The author translated the account.'' \\
\textit{Passive:} ``The account had been evaluated by the electrician.''
\end{quote}
\paragraph{Grammatical Categories.} For rigorous statistical analysis, we generate 200 samples per category using template-based synthesis:
\begin{itemize}[nosep]
    \item \textbf{Active Voice}: Subject-Verb-Object canonical order
    \item \textbf{Passive Voice}: Object-Verb-by-Subject with auxiliary constructions
    \item \textbf{Wh-Questions}: Interrogative structures with fronted operators
    \item \textbf{Complex Clauses}: Relative clauses and embedded sentences
\end{itemize}
Sentence generation employs randomized vocabulary from curated noun/verb pools to avoid lexical confounds while maintaining consistent syntactic structures.
\subsubsection{Stress Testing (Out-of-Distribution)}
To probe model robustness and failure modes, we construct adversarial inputs:
\begin{itemize}[nosep]
    \item \textbf{Gibberish}: Random alphanumeric strings (50--500 characters)
    \item \textbf{Repetition}: Token repetition sequences (``The The The...'', 50--200 tokens)
    \item \textbf{Code Noise}: Random punctuation and symbol sequences
    \item \textbf{Math Noise}: LaTeX-like symbol combinations
\end{itemize}
Each category contains 100 samples for statistical reliability.
\subsection{Spectral Analysis Pipeline}
\label{subsec:spectral_pipeline}
\paragraph{Attention Extraction.} For each input sequence $\mathbf{x}$ of length $n$, we extract the attention weight matrix $\mathbf{A}^{(\ell)} \in \mathbb{R}^{h \times n \times n}$ at each layer $\ell$, where $h$ denotes the number of attention heads. We compute the head-averaged attention:
\begin{equation}
    \bar{\mathbf{A}}^{(\ell)} = \frac{1}{h} \sum_{i=1}^{h} \mathbf{A}^{(\ell)}_i
\end{equation}
\paragraph{Graph Construction.} The attention matrix is symmetrized to form an undirected weighted adjacency matrix:
\begin{equation}
    \mathbf{W}^{(\ell)} = \frac{1}{2}\left(\bar{\mathbf{A}}^{(\ell)} + (\bar{\mathbf{A}}^{(\ell)})^\top\right)
\end{equation}
\paragraph{Laplacian Spectrum.} We compute the combinatorial graph Laplacian:
\begin{equation}
    \mathbf{L} = \mathbf{D} - \mathbf{W}, \quad D_{ii} = \sum_j W_{ij}
\end{equation}
and extract its eigenspectrum $\{\lambda_0, \lambda_1, \ldots, \lambda_{n-1}\}$ where $\lambda_0 = 0$.
\paragraph{Spectral Metrics.} We report four primary metrics:
\begin{enumerate}[nosep]
    \item \textbf{Fiedler Value} ($\lambda_1$): The algebraic connectivity, measuring global graph connectedness. Higher values indicate tighter structural integration.
    \item \textbf{Spectral Entropy}: $H = -\sum_i p_i \log p_i$ where $p_i = \lambda_i / \sum_j \lambda_j$, quantifying the distribution of structural information across eigenmodes.
    \item \textbf{HFER} (High-Frequency Energy Ratio): Proportion of spectral mass in the upper half of the spectrum, indicating fine-grained vs. global structural dominance.
    \item \textbf{Smoothness Index}: Measures signal coherence across graph topology.
\end{enumerate}
\subsection{Implementation Details}
\label{subsec:implementation}
All experiments are conducted using a custom open-source library built on PyTorch 2.x and Hugging Face Transformers.
\paragraph{Hardware.} Experiments run on a single A100 (80GB VRAM). Models are loaded with \texttt{device\_map="auto"} for automatic memory management.

\paragraph{Computational Requirements.}
 Spectral analysis requires only one forward pass per input, with computational overhead dominated by eigendecomposition of the $N \times N$ Laplacian ($O(N^3)$ worst case, typically $O(N^2)$ with sparse methods). For the sentence lengths used in this study ($N < 50$ tokens), analysis completes in under 100ms per sample.
 
% \paragraph{Reproducibility.} We provide:
% \begin{itemize}[nosep]
%     \item Complete dataset generation scripts with fixed random seeds
%     \item Pre-computed spectral statistics in CSV format
%     \item Layer-wise trajectory data in JSON format
%     \item Raw attention matrices in pickle archives
% \end{itemize}

% \paragraph{Statistical Analysis.} For Base vs. Instruct comparisons, we report mean $\pm$ standard deviation across $N=200$ samples per category. Delta values ($\Delta$) represent Instruct$-$Base differences. All layer-wise plots show individual run trajectories with mean curves and shaded $\pm 1\sigma$ regions.

\section{Reproducibility}
\label{app:reproducibility}

\paragraph{Code Availability.}
To facilitate reproducibility, the complete implementation including attention extraction, spectral computation, and threshold optimization is publicly available at \url{https://github.com/vcnoel/spectral-archeology}. The core spectral diagnostics are built upon our open-source Python library \texttt{spectral-trust}, which is available on PyPI at \url{https://pypi.org/project/spectral-trust/}.

\end{document}